\pdfoutput=1
\documentclass[11pt]{article}

\usepackage[final]{acl}

\usepackage{times}
\usepackage{latexsym}
\usepackage{array}
\usepackage{longtable}
\usepackage{enumitem}
\usepackage{multirow}
\usepackage{xcolor}
\usepackage{amsmath}
\usepackage{amsthm}
\usepackage[T1]{fontenc}
\usepackage[utf8]{inputenc}

\usepackage{microtype}

\usepackage{inconsolata}

\usepackage{graphicx}

\usepackage{booktabs}
\usepackage{float}
\usepackage{pifont}
\usepackage{xcolor}
   \definecolor{darkgreen}{rgb}{0.0, 0.5, 0.0}
   \definecolor{darkred}{rgb}{0.5,0.0,0.0}
\newtheorem{definition}{Definition}

\usepackage{siunitx}
\usepackage{makecell}

\usepackage{stfloats}

\usepackage{tabularx, booktabs}
\newcolumntype{L}{>{\raggedright\arraybackslash}X}

\title{Measuring Risk of Bias in Biomedical Reports: The RoBBR Benchmark} 

\author{%
Jianyou Wang\thanks{Equal contribution} \quad Weili Cao\footnotemark[1] \quad Longtian Bao \quad Youze Zheng \quad Gil Pasternak \\
\textbf{Kaicheng Wang} \quad \textbf{Xiaoyue Wang} \quad \textbf{Ramamohan Paturi} \quad \textbf{Leon Bergen} \\
Laboratory for Emerging Intelligence \\
University of California, San Diego\\
\texttt{\{jiw101, w2cao, rpaturi, lbergen\}@ucsd.edu}\\
}

\begin{document}

\maketitle

\begin{abstract}
Systems that answer questions by reviewing the scientific literature are becoming increasingly feasible. To draw reliable conclusions, these systems should take into account the quality of available evidence from different studies, placing more weight on studies that use a valid methodology. We present a benchmark for measuring the methodological strength of biomedical papers, drawing on the risk-of-bias framework used for systematic reviews. Derived from over 500 biomedical studies, the three benchmark tasks encompass expert reviewers' judgments of studies' research methodologies, including the assessments of risk of bias within these studies. The benchmark contains a human-validated annotation pipeline for fine-grained alignment of reviewers' judgments with research paper sentences. Our analyses show that large language models' reasoning and retrieval capabilities impact their effectiveness with risk-of-bias assessment. The dataset is available at \href{https://github.com/RoBBR-Benchmark/RoBBR}{https://github.com/RoBBR-Benchmark/RoBBR}.

\end{abstract}

\begin{figure*}
    \centering
    \includegraphics[width=\linewidth]{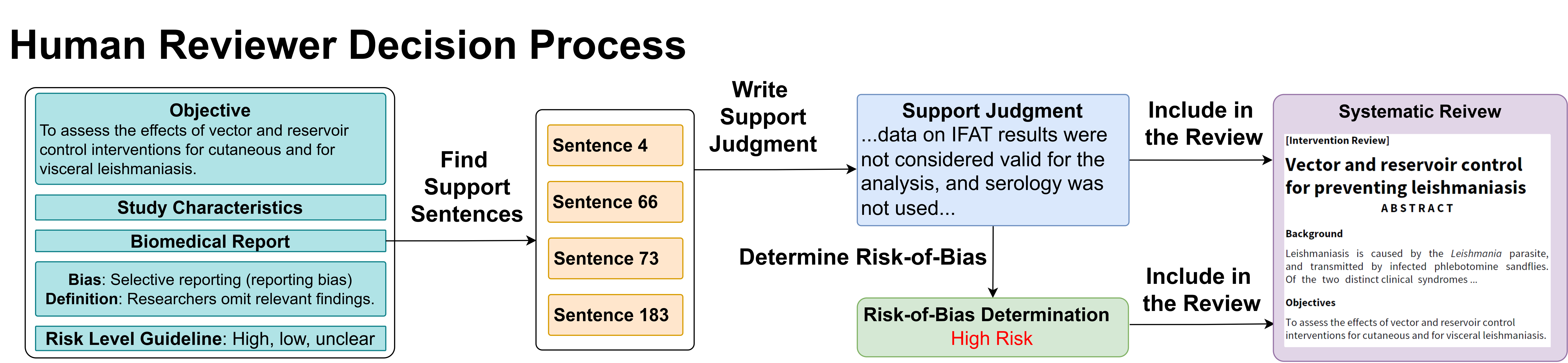}
    \caption{\small{Reviewer finds Sentence 4, 66, 73, 183. The reviewer writes a support judgment based on these sentences. The support judgment indicates this biomedical study has a high risk for selective reporting bias. Both support judgment and the high risk rating are included in the systematic review.
    }}
    \label{fig:flowchart}
\end{figure*}

\section{Introduction}
\label{sec:introduction}

Systems that automatically answer questions by reviewing the scientific literature are becoming increasingly feasible. These systems, such as "Deep Research" offerings from OpenAI, Google, and Anthropic, have the potential to provide scientists with on-demand access to knowledge that is synthesized from across the literature. This can help scientists understand what is already known about topics outside of their research focus, and help clinicians and practitioners keep up to date with best practices. 

 When assessing what is known about a field, not all studies should be weighed equally \cite{cochrane_chp_7}. Studies with stronger methodologies contribute more to a body of evidence than those with weaker methodologies. By weighing studies appropriately, systems can increase the reliability of their summaries and recommendations \cite{bias_adjusted_analysis}. 

 In biomedical research, there is a large body of work which investigates the factors that decrease the validity of a study's methods, and best practices for addressing these issues \cite{cochrane_chp_7, models_for_biased_evidence}. For instance, reporting bias occurs when there are systematic differences in how outcomes are reported or disclosed between the groups that are compared (e.g., the results of the treatment group are more frequently or favorably published than those in the placebo group). This bias can be mitigated by registering the study in advance and committing to publish all results \cite{adressing_reporting_bias}. 
 
 We introduce the RoBBR benchmark for evaluating the methodological strength of biomedical studies, which is referred to as their risk-of-bias levels in the context of this paper. RoBBR's main task involves labeling these risk-of-bias levels. Importantly, we also introduce two novel subtasks: support sentence retrieval (SSR) and support judgment selection (SJS). The SSR subtask is designed to test a model's ability to identify support sentences — specific sentences within an entire paper that signal potential biases. We created the SSR dataset using our annotation pipeline. The SJS subtask evaluates a model's capacity to synthesize this retrieved information through reasoning to form a well-supported judgment. Through our in-depth analysis, we find that the abilities to reason and, especially, to retrieve information are important milestones in improving models' capacity to assess risk-of-bias levels.

\section{Background}
\label{sec:background}
We provide more background on systematic reviews and the risk-of-bias guidelines.

A systematic review is a method in evidence-based medicine that synthesizes evidence from multiple studies to answer important clinical questions. They help researchers and healthcare professionals make informed decisions based on the best available evidence \cite{intro_to_meta_analysis, cochrane_chp_1, cochrane_chp_10}.

 Risk-of-bias assessment involves evaluating each study to determine the likelihood that its results may be biased \cite{cochrane_chp_8, epoc, acrobat}. Studies with a high risk of bias may overestimate or underestimate the true treatment effect, leading to inaccurate conclusions in the systematic review.

 RoBBR follows the risk-of-bias guideline developed by Cochrane \cite{rob2, cochrane_chp_8, cochrane_tool_rob, epoc, acrobat}, a global independent network that conducts systematic reviews of healthcare interventions. Cochrane's risk-of-bias guideline is widely recognized as a standard for evaluating the quality and reliability of research studies. It serves as an important resource for national health agencies in various countries, and are used to formulate national and international guidelines on healthcare practices \cite{US, UK, Australia, health_guideline}. 
 
 During reviewer's assessment of a study's risk of bias, two review authors independently evaluate the risk of bias using the Cochrane tool and resolve disagreements through discussion. They record their rationale for each bias assessment. These rationales are referred to as "support judgments".

\begin{figure*}
    \centering
    \includegraphics[width=\linewidth]{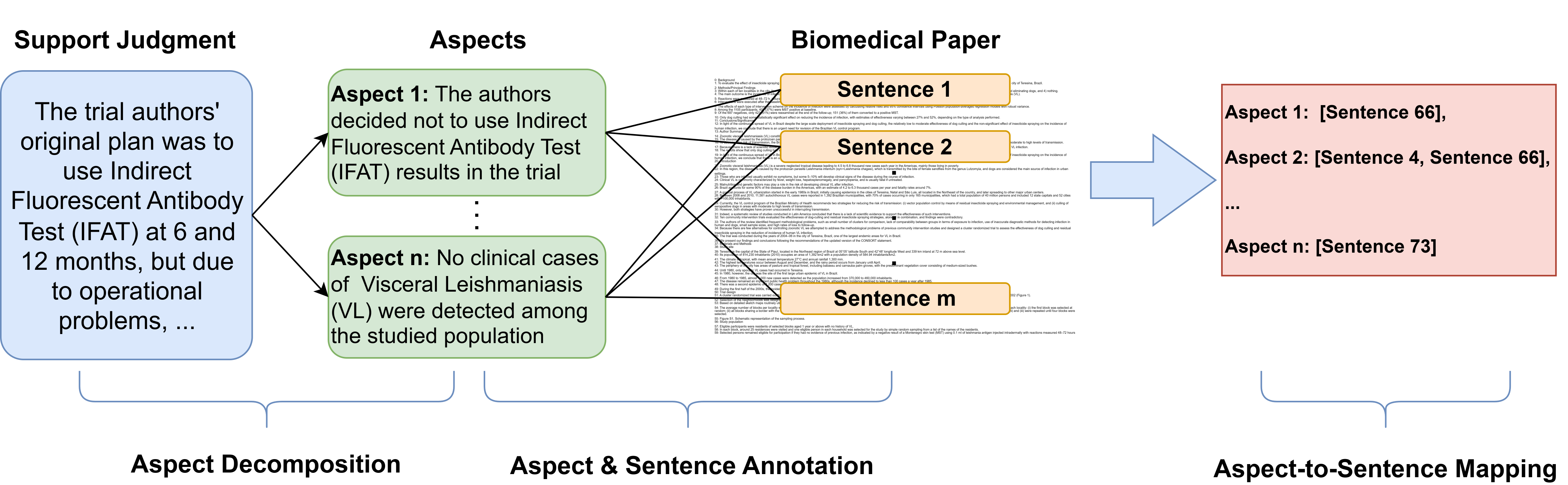}
    \caption{\small{Given a support judgment, our annotation pipeline produces the Aspect-to-Sentence Mapping, a many-to-many relationship. The mapping on the right-side shows Sentence 66 covers both Aspect 1 and 2, so if a model retrieves Sentence 66, it should not retrieve Sentence 4 to avoid redundancy.}}
    \label{fig:gpt_tracer}
\end{figure*}

\section{Related Work}
\label{sec:related work}

{\bf Risk-of-Bias Labeling\\}
Previous research focuses on creating datasets and training models to label risk-of-bias levels \cite{llm_rob_1, llm_rob_2, rob_preclinical_nlp, rob_ml_old}. Traditionally, machine learning models including support vector machines \cite{svm_1, robotreviewer_acl, svm_2}, convolutional neural networks \cite{robotreviewer_acl, cnn} and logistic regression \cite{LR_1, LR_2} are used for label prediction. More recently, some work employs transformers \cite{robin} and LLMs to determine risk-of-bias levels \cite{llm_auto_1, llm_auto_2, llm_auto_3}. While RoBBR's main task is similar to existing work, its novel contribution is the introduction of two new subtasks: support sentence retrieval (SSR) and support judgment selection (SJS). These subtasks identify crucial intermediate steps in risk-of-bias labeling and respectively measure the information retrieval and reasoning capabilities of LLMs when they assess risk-of-bias.

\noindent{\bf Support Sentences in Risk-of-Bias Labeling\\}
Previous studies \cite{robotreviewer, robin} attempt to extract supporting sentences from reviewers' support judgments using direct quotes \cite{robotreviewer} or semantic embeddings \cite{robin}. However, these techniques falter because support judgments are free-form and frequently omit direct quotes, while semantic embeddings are imprecise and incomprehensive (See Table \ref{tab:hypothesis_test}). This makes sentences extracted by such methods unsuitable for a rigorous benchmark like our Support Sentence Retrieval (SSR) subtask, which tests an LLM's capability in retrieving support sentences.  To address this, in Section \ref{ssec:gpt tracer}, our annotation pipeline aligns all aspects (such as paraphrases, synthesized and summarized information) from support judgment to sentences in biomedical studies, which would more reliably and comprehensively extract support sentences that are used to form the novel SSR subtask. 

\noindent{\bf LLM Extraction from Systematic Reviews}\\
In recent years, LLMs have been used to extract data from systematic reviews, such as experiment evidence \cite{llm_auto_data_extract, llm_extract}, summaries \cite{md_summary_lit_reviews, rct_summary}, quality of evidence \cite{automating_quality_assessment} and PICO \cite{pico_extraction_gpts}. In our work, aligning the human support judgment to various sentences in biomedical studies is a more complex and difficult "data extraction" task, as noted by \citeauthor{robin}, which necessitates the aspect decomposition and aspect \& sentence annotation components in our annotation pipeline. 

\section{Benchmark Development}
\label{sec:dataset description}

\subsection{Benchmark Statistics and Data Sources}
\begin{table}[htbp]
    
    \centering
    \setlength{\tabcolsep}{2pt}
    \small
    \begin{tabular}{lcccccc}
        \toprule
        & \textbf{Cochrane} & \textbf{Cochrane} & \textbf{Non-Cochrane}\\
        \textbf{Task} & \textbf{Train} & \textbf{Test} & \textbf{Test}\\
        \midrule
        Retrieval (SSR) & n=235 & n=313 & N/A\\
        Selection (SJS) & n=346 & n=465 & N/A\\
        Risk-of-Bias (Main) & n=774 & n=906 & n=2,489\\
        \bottomrule
    \end{tabular}
    \caption{{\small RoBBR Benchmark Statistics}.}
    \label{tab:testset_statistics}
\end{table}

 There are two prevailing risk-of-bias assessment guidelines, RoB1 \cite{cochrane_handbook_5_1} and the updated RoB2 \cite{rob2}. As of 2025, many recently published and updated systematic reviews still use RoB1, for example, \cite{rob1_example1, rob1_example2}. Since the RoB2 guidelines do not require support judgments that are necessary for our subtasks, our subtasks only include systematic reviews that follow RoB1. For the main task, we include systematic reviews that follow either set of guidelines. 
 
 Since there are hundreds of bias names and definitions, such as \textit{"incomplete adverse event reporting (reporting bias)"} and \textit{"selective outcome reporting (reporting bias)"},
 for clarity of presentation, we manually group them into 6 primary categories: "selection", "attrition", "performance", "detection", "reporting" and "deviation" (unique to RoB2) using two approaches. In some cases, bias names already include broader category labels (e.g., "random sequence generation (selection bias)"). When this happens, we follow these broader category labels. When bias names lack such broader category labels, we manually identify an equivalent bias name that has a category label and use this label. For example, the bias name "bias arising from the randomization process" is equivalent to "random sequence generation (selection bias)," so we use "selection bias" as its category label. A bias name belongs to multiple categories when these categories are explicitly included in the bias name. For example, a bias called "blinding for adverse event (performance and detection bias)" is categorized into both the "performance bias" category and the "detection bias" category. See Table \ref{tab:bias-categorization} for examples of how bias names are categorized.
 
During evaluation, models would see the actual bias name and definition used in systematic reviews. Bias definitions are sourced from systematic reviews themselves and official guidelines like \cite{cochrane_handbook_5_1, epoc, acrobat}. 

 For data diversity, we include both systematic reviews that are published in Cochrane (denoted as Cochrane Train/Test) and that are published elsewhere (denoted as Non-Cochrane Test\footnote{\small{Empirically, Llama3 fine-tuned on Cochrane train set also perform well on Non-Cochrane test set, so we did not create a separate Non-Cochrane train set.}}). Non-Cochrane reviews do not have support judgments and cannot form our subtasks.

RoBBR includes 58 Cochrane reviews that assess 204 papers, and 496 Non-Cochrane reviews that assess 496 papers. Appendix \ref{ssec:appendix_dataset statistics} show details on token length, bias and label distributions. See Table \ref{tab:testset_statistics} for statistics of the three tasks in our benchmark.

\subsection{Main Task: Risk-of-Bias Determination}
\label{ssec: risk of bias: risk level determination}
\begin{figure*}[ht]
    \centering
    \includegraphics[width=\linewidth]{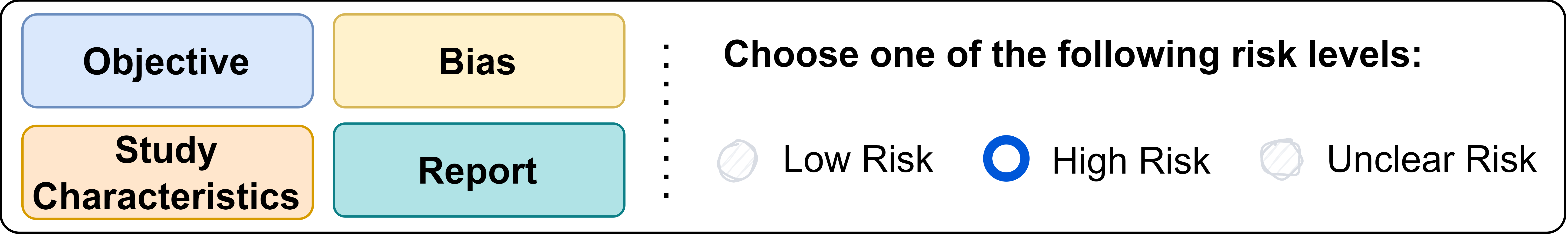}
    \caption{Main Task: Risk-of-Bias Determination. The goal of the main task is to provide an assessment of the paper's risk of specific biases.}
    \label{fig:task4}
\end{figure*}

The main task evaluates a model's ability to label the risk-of-bias of a biomedical study as high, low, or unclear/some concern. The input to the model is exactly everything that a human expert would see, including the biomedical paper, study characteristics (i.e. PICO), the objective/topic of the systematic review, one specific bias name and definition, and risk level definition. See the top-left panel of Figure \ref{fig:flowchart} for a visualization of input. The same input would be used for the two subtasks as described in section \ref{ssec:task 2: risk of bias sentence retrieval}, \ref{ssec:task 3: risk of bias support judgment selection}. To visualize the task, see Figure \ref{fig:task4}.

The decision-making process of risk-of-bias determination is highly complex and involves multiple intermediate steps: retrieve support sentences from the biomedical paper, synthesize evidence from support sentences into a support judgment, and follow the guidelines and definitions of bias and risk level to give a final determination.

\subsection{Automatic Annotation Pipeline}
\label{ssec:gpt tracer}
Support judgments contain a wealth of information, including evidence, data, reasoning, and commentaries written by expert reviewers. The goal of support judgment is to explain why reviewers would give this risk-of-bias rating. Therefore, support judgments mirror and shed light on the decision-making process of the reviewer. 

 It is difficult to find all support sentences in the study that form the basis of a support judgment. As shown in Figure \ref{fig:gpt_tracer}, support judgments can have many aspects. While some aspects are direct quotes from the study, many other aspects are paraphrases from the study, or deeper analyses that synthesize information from multiple parts of the study.

 To reverse engineer a support judgment, we find that no lexicon-based or embedding models can find all support sentences accurately (see Section \ref{ssec:evaluating llm's ability against human annotators} and Table \ref{tab:hypothesis_test}). To address this, our proposed annotation pipeline has two main components: Aspect Decomposition and Aspect \& Sentence Annotation. See Figure \ref{fig:gpt_tracer} for a visualization. GPT-4 (gpt-4-0125) is used by our annotation pipeline. All prompts used in our annotation pipeline are in Appendix \ref{sec:appendix_task 2 prompt optimization}.

\noindent\textbf{Aspect Decomposition:}\\
To reduce the complexity of the reverse engineering process, we decompose the support judgment into distinct, non-overlapping pieces of information, each focusing on a specific aspect of the original support judgment. This is accomplished by careful prompt engineering with GPT-4. We manually validated that all decomposed aspects from all support judgments are of high quality.

\noindent\textbf{Aspect \& Sentence Annotation:}\\
To further reduce complexity and instability introduced by long context input to LLMs, we only ask GPT-4 to determine if one sentence from the study covers one aspect.

\begin{definition}
A sentence $s_j$ in a paper \textbf{covers} an aspect $f_i$ if it satisfies these two criteria:\\
1: The content of the sentence conveys the majority of the aspect's information.\\
2: Any information about the aspect not directly stated in the sentence can be reasonably inferred from the surrounding context.
\end{definition}

 We aggregate these annotations into the \textbf{Aspect-to-Sentence Mapping:}
For each (study, support judgment) triplet, a sentence is mapped to several aspects (e.g., 0, 1, 2, or more). Most sentences are not mapped to any aspect because they do not cover them. A few sentences can even be mapped to more than one aspect because aspect-to-sentence mapping is a many-to-many relationship. 

 The total cost of our annotation pipeline via OpenAI API is around \$1,000 with aspect decomposition and initial filtering costing less than \$100. Next, we evaluate the quality of GPT-4 annotation.

\begin{table}[tbp]
    \centering
    \scriptsize
    \setlength{\tabcolsep}{4pt}
    \begin{tabular}{lccc}
        \toprule
        \textbf{Metrics} & \textbf{Human \& Human} & \textbf{Human \& GPT} & \textbf{p-value} \\
        \midrule
        Exact Accuracy   & 99.4 $\pm$ 0.2 & 99.3 $\pm$ 0.2 & 0.31  \\
        F1 Binary        & 73.4 $\pm$ 6.3 & 71.7 $\pm$ 5.6 & 0.55  \\
        Cohen's $\kappa$ & 73.1 $\pm$ 6.3 & 71.4 $\pm$ 5.6 & 0.54  \\
        Spearman's $\rho$& 73.8 $\pm$ 6.0 & 71.7 $\pm$ 5.6 & 0.43  \\
        \midrule
        \textbf{Metrics} & \textbf{Human \& Human} & \textbf{Human \& Embedding} & \textbf{p-value} \\
        \midrule
        Exact Accuracy   & 99.4 $\pm$ 0.2 & 98.5 $\pm$ 0.2 & $<\!10e{-3}$  \\
        F1 Binary        & 73.4 $\pm$ 6.3 & 30.5 $\pm$ 8.1 & $<\!10e{-3}$  \\
        Cohen's $\kappa$ & 73.1 $\pm$ 6.3 & 29.7 $\pm$ 8.2 & $<\!10e{-3}$  \\
        Spearman's $\rho$& 73.8 $\pm$ 6.0 & 29.8 $\pm$ 8.3 & $<\!10e{-3}$  \\
        \bottomrule
    \end{tabular}
    \caption{{\small Inter-annotator agreement. \textbf{Top:} comparison between two human teams and GPT-4. \textbf{Bottom:} comparison between two human teams and OpenAI-v3-large as a post-hoc analysis}.}
    \label{tab:hypothesis_test}
\end{table}

\subsubsection{Evaluating Annotation Quality}
\label{ssec:evaluating llm's ability against human annotators}
We hypothesize a sentence matches an aspect is straightforward enough for GPT-4 to achieve human-level performance with a specialized prompt. Following the protocol introduced by \cite{dorismae}, we optimized the instruction prompt for GPT-4 on a development set (see Appendix \ref{sec:appendix_task 2 prompt optimization}). 
We randomly sampled 50 papers, with each assigned a single aspect. Across all 50 papers, this resulted in a total of 13,575 (aspect, sentence) pairs. Four graduate student were tasked with annotating these pairs. The annotators were divided into two teams, with each team consisting of two annotators. The annotators' training for annotation is described in Appendix \ref{ssec:appendix Annotator Training}. Each annotator performed the annotation tasks individually given a pre-defined annotation guideline in Appendix \ref{sec:appendix_annotation guideline}. Annotators from the same team then collaborated to resolve differences and eliminate mistakes. 

 Each team produced a set of annotation results. GPT-4-0125 annotated the same 13k (aspect, sentence) triplets. We calculated Exact Accuracy, F1-binary, Spearman Correlation, and Kappa Coefficient between the two human teams and between each human team and GPT-4. Table \ref{tab:hypothesis_test} shows human-human correlations in the low-to-mid 70s, indicating reasonable agreement, despite the fact that finding sentences that cover aspects is non-trivial due to different annotators' potentially varied interpretations of key terms in \textbf{Definition 1} such as \textit{"majority"} and \textit{"reasonably inferred"}. For context on inter-annotator agreement in similar aspect-based tasks, \citeauthor{dorismae} reported 54\% $\rho$ (three-way annotation) and \citeauthor{csfcube} found near 40\% pre-adjudication $\rho$ (four-way annotation). 

 Table \ref{tab:hypothesis_test} shows human-GPT4 and human-human correlations are comparable. A post-hoc analysis reveals that powerful embedding models like OpenAI-v3-large \cite{openai_text_embedding_3} have significantly lower correlation with humans. For the primary human-GPT4 comparison, the p-value (p = 0.3) from our bootstrapped test fails to reject the null hypothesis (no difference in correlation). In contrast, we can reject the null hypothesis for the embedding model which has clear and detectable differences from human annotators. 

\subsection{Support Sentence Retrieval (SSR)}
\label{ssec:task 2: risk of bias sentence retrieval}
With our annotation pipeline, we transform a support judgment into a special information retrieval task that evaluates if a model can identify and retrieve these support sentences from the full text of the paper. Since these support sentences form the basis of the support judgment, if a model can locate these support sentences, it is one step closer to reaching a good support judgment and classifying the correct risk-of-bias level. 

Figure/table captions are split into sentences as text. Tables are included and turned into markdown format and treated as individual sentences. Figures are excluded since SSR is designed to evaluate textual models. 

 We have obtained the \textbf{Aspect-to-Sentence Mapping}. From this mapping, there exists the smallest optimal integer $K$ such that there are $K$ sentences that can cover all aspects in a support judgment. Therefore, when a model is only allowed to retrieve $K$ sentences from a study, we measure the percentage of aspects that the model's retrieved set of sentences can cover. This metric is denoted as \textbf{Aspect Recall Ratio @ Optimal}.

 Formally, let $\{s_1, \dots, s_k\}$ be the set of retrieved sentences, and $\{f_1, \dots, f_m\}$ be the set of aspects. We define the indicator function $\mathcal{S}(f_i, s_j)$ as:
\[
\mathcal{S}(f_i, s_j) = 
\begin{cases} 
1 & \text{if } s_j\text{ covers } f_i \\
0 & \text{otherwise}
\end{cases}
\]

 The \textbf{Aspect Recall Ratio @ Optimal} is defined as

\begin{equation}
\frac{\sum_{i=1}^{m} \displaystyle \mathbf{1}\Bigg\{{\left(\sum_{j=1}^{K} \mathcal{S}(f_i, s_j)\right) \geq 1}\Bigg\}}{m}
\label{eq:aspect_recall}
\end{equation}

 Not only does it evaluate if a model can find support sentences, but it also evaluates if a model can control information redundancy from these sentences. Figure \ref{fig:task2} in Appendix \ref{sec:appendix task visual} visualizes SSR. 

\begin{table*}[htbp]
\centering
\begin{minipage}{0.48\textwidth}
    \scriptsize
    \setlength{\tabcolsep}{1.8pt}
    \begin{tabular}{l|cccccc}
    \toprule
    & \multicolumn{6}{c}{\textbf{Bias Type}}\\
        Model & Avg & Selection & Attrition & Performance & Detection & Reporting\\
    &  & n = 157 & n = 46 & n = 54 & n = 61 & n = 14 \\
\midrule
OpenAI-v3        & 22.7 & 40.1 & 6.3 & 26.7 & 27.4 & 13.1\\
GritLM-7B        & 18.9 & 35.8 & 7.0 & 15.2 & 26.8 & 9.5\\
\midrule
% OpenAI o3           & \textbf{53.2} & \textbf{70.0} & 38.3 & \textbf{59.8} & \textbf{54.7} & \textbf{43.0}\\
GPT-4o           & \textbf{47.5} & 60.5 & \textbf{42.4} & 41.5 & \textbf{50.1} & \textbf{43.0}\\
Sonnet-3.5       & 39.2 & 55.4 & 30.7 & 37.0 & 37.2 & 35.5\\
Llama-3.1-70B    & 45.6 & 61.2 & 34.2 & 45.2 & \textbf{50.1} & 37.4\\
Llama-3-8B       & 22.7 & 49.4 & 14.5 & 22.1 & 20.8 & 6.5\\
\midrule
Llama-3-8B       & \multirow{2}{*}{40.8} & \multirow{2}{*}{\textbf{69.0}} & \multirow{2}{*}{24.8} & \multirow{2}{*}{\textbf{47.3}} & \multirow{2}{*}{48.6} & \multirow{2}{*}{14.3}\\
Fine-tuned     &      &      &      &      &      &\\
\bottomrule
\end{tabular}
\caption{\small{Support Sentence Retrieval (SSR). Evaluated on Cochrane Test (SSR). Metric is Aspect Recall Ratio @ Optimal. Llama-3-8B fine-tuned on Cochrane Train (SSR). Fine-tuning details are included in Appendix \ref{appendix:fine-tuning}.}}
\label{tab:task_2_sentence_retrieval_results}
\end{minipage}
\hspace{0.02\textwidth}
\begin{minipage}{0.48\textwidth}

\renewcommand{\arraystretch}{1.3}
\centering
\scriptsize
\setlength{\tabcolsep}{1.8pt}
\begin{tabular}{l|cccccc}
\toprule
    & \multicolumn{6}{c}{\textbf{Bias Type}}\\
    Model & Avg & Selection & Attrition & Performance & Detection & Reporting\\
    &  & n = 130 & n = 98 & n = 87 & n = 82 & n = 80\\
\midrule
% OpenAI o3           & \textbf{60.8} & \textbf{73.1} & 70.4 & 48.3 & \textbf{56.1} & \textbf{56.3}\\
GPT-4o           & 47.2 & 58.5 & 60.2 & 48.3 & 42.7 & 26.3\\
Sonnet-3.5       & \textbf{59.9} & \textbf{73.1} & \textbf{73.5} & \textbf{50.6} & \textbf{51.2} & \textbf{51.3}\\
Llama-3.1-70B    & 53.2 & 66.2 & 62.2 & 44.8 & 46.3 & 46.3\\
Llama-3-8B       & 26.5 & 26.9 & 34.7 & 24.1 & 22.0 & 25.0\\
\midrule
Llama-3-8B       & \multirow{2}{*}{29.6} & \multirow{2}{*}{40.8} & \multirow{2}{*}{28.6} & \multirow{2}{*}{24.1} & \multirow{2}{*}{19.5} & \multirow{2}{*}{35.0}\\
Fine-tuned       &      &      &      &      &      &\\
\bottomrule
\end{tabular}
\caption{\small{Support Judgment Selection (SJS). Evaluated on Cochrane Test (SJS). Metric is Accuracy. Llama-3-8B fine-tuned on Cochrane Train (SJS). Fine-tuning details are included in Appendix \ref{appendix:fine-tuning}}.}
\label{tab:task_3_support_judgment_selection_results}
\end{minipage}
\end{table*}

\begin{table*}[htbp]
    \centering
    \scriptsize
    \setlength{\tabcolsep}{12.5pt}
    
    \begin{tabular}{l|ccccccc}
    \toprule
    & \multicolumn{7}{c}{\textbf{Bias Type}}\\
    
    Model & Avg & Selection & Attrition & Performance & Detection & Reporting & Deviation\\ 
    & &  n = 933 & n = 629 & n = 309 & n = 645 & n = 594 & n = 331\\

\midrule
GPT-4o& \textbf{42.1} & 50.5 & 36.5 & \textbf{54.7} & 43.7 & \textbf{34.4} & 32.7\\ 
Sonnet-3.5& 41.9 & \textbf{52.8} & \textbf{37.1} & 50.6 & 43.3 & 33.6 & \textbf{34.2}\\
Llama-3.1-70B & 38.8 & 48.1 & 31.7 & 48.0 & \textbf{44.4} & 31.6 & 29.0 \\
Llama-3-8B & 30.1 & 36.4 & 32.4 & 39.1 & 37.2 & 19.8 & 15.4  \\
\midrule
Llama-3-8B Fine-tuned& 36.3  & 49.5 & 33.3 & 42.4 & 40.1 & 27.0& 25.7 \\
LR Trained & N/A & 26.0 & 21.7 & 27.9 & 28.6 & 21.8 & N/A \\
SVM Trained & N/A & 30.7 & 24.2 & 23.3 & 25.6 & 23.5 & N/A \\
    \bottomrule
    \end{tabular}
\caption{{\small Main Task, Risk-of-Bias Determination. Evaluated on Cochrane Test (Main) + Non-Cochrane Test (Main). Metric is Macro-F1. Llama-3-8B, logistic regression, and SVM are fine-tuned/trained on Cochrane Train (Main). Fine-tuning details are included in Appendix \ref{appendix:fine-tuning}}.}
\label{tab:task_4_risk_level_determination_results}
\end{table*}

\subsection{Support Judgment Selection (SJS)}
\label{ssec:task 3: risk of bias support judgment selection}
 Simply retrieving support sentences is insufficient for a risk-of-bias determination. A model needs the ability to interpret and synthesize information from these retrieved support sentences in order to generate a good support judgment that will lead to the correct determination of risk-of-bias. 

 Following the convention of using multiple-choice questions as a proxy for generative tasks \cite{gpqa}, we propose the support judgment selection subtask, an MCQ task, where the model selects the correct support judgment from a mix of three synthetically generated support judgments, and three human-written options derived from other papers' support judgments for the same exact bias. These three human-written options sometimes proved to be surprisingly hard distractors because their lack of direct paper quotes made them into generic, non-paper-specific statements. A deep logical understanding of the paper's content rather than superficial semantic matching is required to look past them and find the correct one.
 
 We prompt GPT-4 to generate the three synthetic options by imitating support judgments from other papers concerning the same exact bias name. These options are tailored to be paper-specific while maintaining the underlying reasoning. Empirically, these six distractor options are often misleading. Table \ref{tab:task_3_support_judgment_selection_results} shows the best LLM only achieves 60\% accuracy.  In order to ensure that all incorrect options were actually incorrect, all 906 datapoints were checked by human, and 465 datapoints were kept. See Figure \ref{fig:task3} in Appendix \ref{sec:appendix task visual} to visualize SJS.

\section{Experiments}
\label{sec:experiments}

\subsection{Evaluated Models}
\label{ssec:experimental procedure}
We evaluate various LLMs, GPT-4o-2024-05-13 \cite{gpt_4o}, Sonnet3.5-20240620 \cite{claude3.5_sonnet}, and Llama-3.1-70B \cite{llama3}. We evaluate and fine-tune light-weight Llama-3-8B \cite{llama3}. We also evaluate two recent embedding models GritLM 7B \cite{gritLM} and OpenAI-embedding-v3-large \cite{openai_text_embedding_3}. Note the GPT-4o model is different from the GPT-4-0125 model used in our annotation pipeline.

For the main task, we further evaluate three traditional ML models: RobotReviewer \cite{robotreviewer}, Support Vector Machine (SVM) and Logistic Regression (LR) \cite{robin}. 

For all generative language models, we use chain-of-thought as a prompting strategy to stabilize the model performance. We develop our prompts on train sets. For reproducibility, we use a common generation setting of temperature 0, top-p 1, frequency penalty 0, and presence penalty 0.

\subsection{Evaluation Metrics}
\label{ssec:evaluation metric}
Some bias names may fall into multiple categories, causing the total datapoint count across categories to be slightly higher than the test set size. Model performance is reported per category and as an average across all categories.

The SSR subtask uses \textbf{Aspect Recall Ratio @ Optimal} defined in Equation \ref{eq:aspect_recall} as its metric. It measures the percentage of aspects covered by the retrieved sentences. Unlike traditional metrics (e.g., recall, precision), it penalizes redundancy by rewarding models that retrieve the minimum number of sentences (typically 1-2) needed to cover all aspects for a given data point. For instance, if one sentence covers multiple aspects, retrieving additional sentences that cover already addressed aspects is discouraged.

\section{Benchmarking Results and Analyses}
\label{sec:results}

\subsection{LLMs Exhibit Distinct Strengths in Retrieval and Reasoning}
\label{ssec:llms have different struggles}

While GPT-4o and Sonnet-3.5, with 42.1\% and 41.9\% average Macro-F1, are comparable in the main task Table \ref{tab:task_4_risk_level_determination_results}, a closer inspection of their performances on the two subtask Tables \ref{tab:task_2_sentence_retrieval_results}, \ref{tab:task_3_support_judgment_selection_results} reveals interesting differences. In support sentence retrieval (SSR), GPT-4o has 47.5\% Aspect Recall Ratio, while Sonnet-3.5 only has 39.2\%. This shows that Sonnet-3.5 is worse at retrieving the best support sentences that are the bases of human expert-level support judgments. However, in support judgment selection (SJS), Sonnet-3.5 outperforms other GPT-4o with 59.9\%  over 47.2\% accuracy. This shows that Sonnet-3.5 can select better support judgments, which involves reasoning and deliberating which option best aligns with the paper's contents and the bias and risk-level definitions. 

In other words, GPT-4o's advantage over Sonnet-3.5 is information retrieval, while Sonnet-3.5's advantage is information synthesis with reasoning. We provide an example in Appendix \ref{sec:appendix_error_analysis} comparing reasoning traces of the two models to clearly illustrate Sonnet-3.5's advantage in reasoning.

\subsection{Positive Correlation Between Main Task and Subtasks}

\begin{figure}[tbp]
    \centering
    \includegraphics[width=\linewidth]{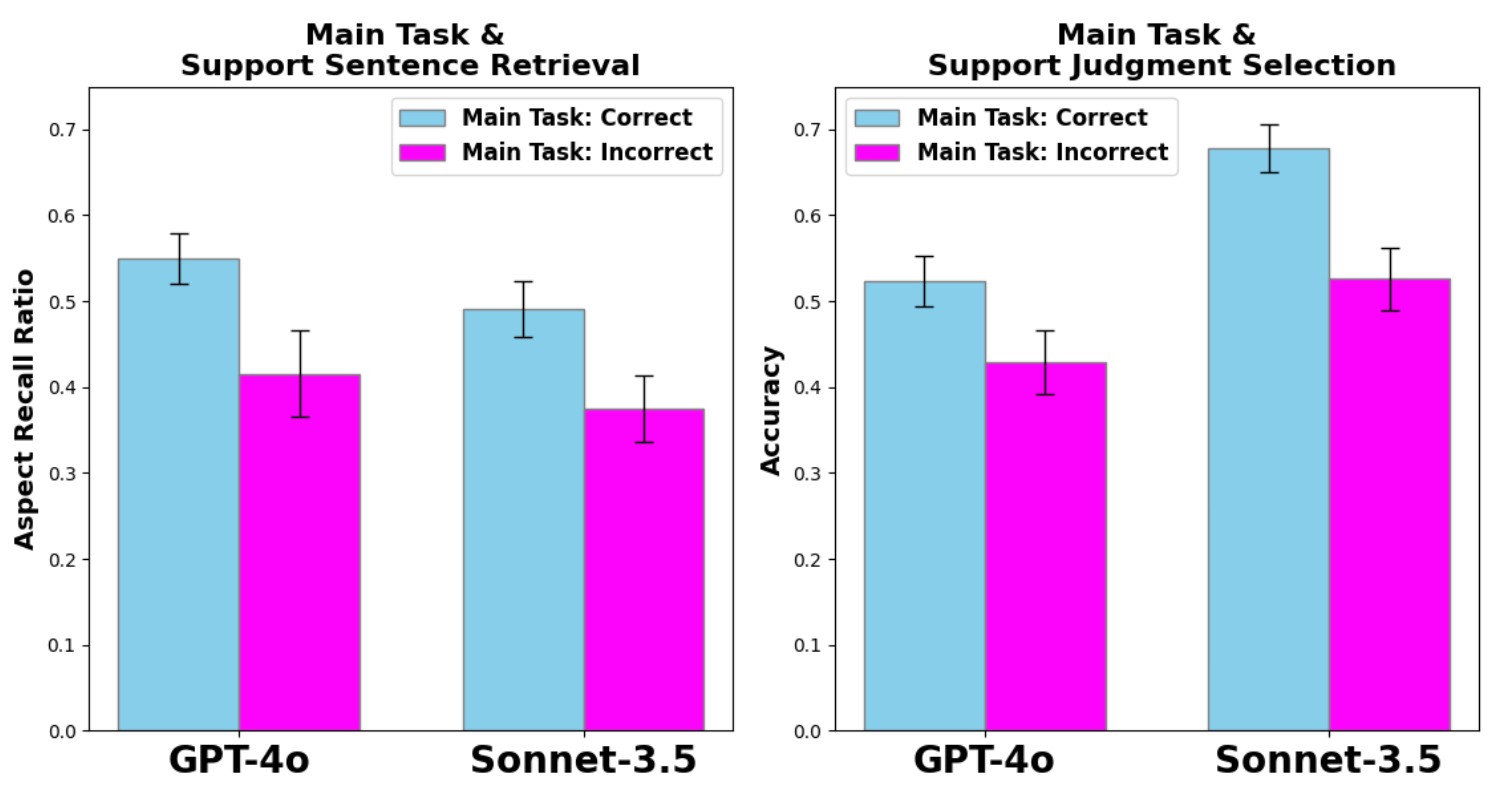}
    \caption{{\small There is a positive correlation between the main task and SSR \& SJS subtask performance.}}
    \label{fig:correlation}
\end{figure}

We investigate the relationship between the main task and our subtasks. For an LLM (either GPT-4o or Sonnet 3.5), we categorize datapoints in the SSR subtask into two classes: datapoints whose risk-of-bias level is correctly predicted by the LLM and those incorrectly predicted. For each of the two classes, we calculate its average Aspect Recall Ratio @ Optimal. See the left part in Figure \ref{fig:correlation}, we observe that the Aspect Recall Ratio is significantly higher in the correct class than the incorrect class, for both Sonnet-3.5 and GPT-4o, suggesting LLMs' performances on SSR are positively correlated with their performances on the main task. 

Similarly, datapoints in the SJS subtask are divided into two classes. From the right part in Figure \ref{fig:correlation}, we see that, for both Sonnet-3.5 and GPT-4o, their performances on SJS are positively correlated with their performances on the main task. 

\subsection{Retrieval and Reasoning Ablation}

\begin{table}[tbp]
\centering
\small
\begin{tabular}{@{}llc@{}}
\toprule
\textbf{Retrieval} & \textbf{Reasoning} & \textbf{Main Task Result} \\
\midrule
Sonnet-3.5   & GPT-4o  & 31.5 $\pm$ 5.0 \\
GPT-4o       & GPT-4o  & 40.3 $\pm$ 6.9 \\
Ground Truth & GPT-4o  & 46.1 $\pm$ 7.5 \\
Ground Truth & Sonnet-3.5   & 49.4 $\pm$ 8.0 \\
\bottomrule
\end{tabular}
\caption{\small Two stage pipeline that first retrieves and then reason based on the retrieved results (full paper is not shown). Main Task, Risk-of-Bias Determination. Evaluated on the 313 datapoints in Cochrane Test (SSR), all of which have risk-of-bias labels. Metric is Macro F1 score (\%) averaged across bias categories. The error bar represents the 95\% confidence interval.}
\label{tab:subtask_ablation}
\end{table}
To investigate how the retrieval and reasoning abilities of LLMs affect their performance on the main task, we employ a two-stage retrieval and reasoning pipeline. In this setup, one LLM processes sentences retrieved by another LLM, rather than accessing the original paper directly. The specific prompt used is detailed in Appendix \ref{ssec:appendix_prompts for main with retrieval}. This experiment was conducted on Cochrane Test (SSR), which comprises 313 datapoints, each associated with a support judgment and a risk-of-bias label.

We pair different models' retrieved sentences with different reasoning models to evaluate performance on the main task Risk-of-Bias Determination. Ground truth retrieved sentences are provided by our annotation pipeline. 

As shown in Table \ref{tab:subtask_ablation}, when results from different retrieval methods are paired with the same reasoning model (GPT-4o), ground-truth retrieval achieves significantly higher macro F1 score (46.1\%) over Sonnet-3.5's retrieval (31.5\%), confirming that retrieval quality impacts main task performance. When the retrieval quality is held constant (using ground-truth retrieval) and the reasoning model varied, there is some inconclusive evidence that suggests Sonnet-3.5 slightly outperforms GPT-4o. These ablations demonstrate that both retrieval and reasoning abilities are relevant for the Risk-of-Bias Determination task. 

\subsection{Retrieval Marginally Improves LLM on Non-Cochrane Test Set}
So far we have focused on the Cochrane test set, which includes support judgments, unlike the Non-Cochrane test set. To investigate whether more accurate support sentences could enhance LLM performance on the Non-Cochrane test set (2489 datapoints), we used the o4-mini reasoning model \cite{openai2024o3} to find such sentences within each paper, guided by the datapoint's actual risk level (see Appendix \ref{ssec:appendix_prompts o4 mini} for the prompt). In this proof-of-concept setup, the LLM received the full paper and o4-mini-identified support sentences as highlights, which we compared against baselines with full paper but no highlights.

Table \ref{tab:retrieval_effect} reveals that both GPT-4o and Sonnet-3.5 showed slight performance improvements on the Non-Cochrane Test (Main) over their baselines. The limited gain is partly due to inaccuracies in retrieved sentences, as recovering support sentences without expert support judgments is challenging.

\begin{table}[tbp]
\centering
\small
\begin{tabular}{@{}lc@{}}
\toprule
\textbf{Setting} & \textbf{Main Task} \\
\midrule
GPT-4o Full Paper & 42.5 $\pm$ 2.2 \\
GPT-4o Full Paper + Highlight Retrieval & 44.1 $\pm$ 2.2 \\
\midrule
Sonnet-3.5 Full Paper & 40.3 $\pm$ 2.2\\
Sonnet-3.5 Full Paper + Highlight Retrieval & 44.5 $\pm$ 2.2 \\
\bottomrule
\end{tabular}
\caption{\small Main Task, Risk-of-Bias Determination. Evaluated on Non-Cochrane Test (Main). Metric is Macro F1 score (\%) averaged across bias categories. The error bar represents the 95\% confidence interval.}
\label{tab:retrieval_effect}
\end{table}

\subsection{Challenges in Support Sentence Retrieval}

Our findings establish the retrieval task SSR as a crucial intermediate milestone for improving LLM on the main task. Nevertheless, Table \ref{tab:task_2_sentence_retrieval_results} indicates that in zero-shot setting, current models struggle significantly: modern embedding models such as OpenAI-v3 and GritLM-7B achieve low recall ($\leq$22.7\%), while commercial LLMs also underperform (recall $\leq$47.5\%). Although fine-tuning LLama3-8B boosts its SSR recall from 22.7\% to 40.8\%, its retrieval quality remains unsatisfactory.

A significant limitation of embedding models is their lack of context awareness. Even recent contextual-aware models \cite{contextual_embedding} cannot handle the long-range interactions across different sentences. In Appendix \ref{sec: appendix long context local vs global}, we examine a type of globally-connected datapoints in SSR potentially necessitates a more global understanding of the entire context. We leave the exploration of contextual embedding models and more accurate long-context LLMs for solving SSR as a future research direction.

\subsection{Limitations of Traditional ML Models on RoBBR}

\begin{table}[htbp]
    \centering
    \scriptsize
    \setlength{\tabcolsep}{6pt}
    \begin{tabular}{l|ccccc}
    \toprule
    & \multicolumn{5}{c}{\textbf{Biases}}\\
    \cmidrule{2-6}
    Model & Avg & Allocate & Blind & Blind & RanSeq  \\ 
    &  & Conceal & Outc & Part & Gen \\
    &  & n=32 & n=19 & n=18 & n=30 \\ 
    \midrule
    RobotReviewer & 56.7 $\pm$ 8.4 & 75.0 & 39.1 & 43.8 & 68.9 \\
    LR              & 53.1 $\pm$ 9.7 & 71.9 & 50.4 & 51.8 & 38.4 \\
    SVM             & 44.8 $\pm$ 8.6 & 45.9 & 55.2 & 41.9 & 36.0 \\
    GPT-4o             & 65.6 $\pm$ 8.5 & 83.6 & 59.1 & 41.9 & 77.8 \\
    Sonnet-3.5          & 67.5 $\pm$ 8.4 & 77.0 & 82.5 & 41.9 & 68.8 \\

    \bottomrule
    \end{tabular}
    
\caption{{\small Main Task, Risk-of-Bias Determination. Evaluated on a subset from the Cochrane Test (Main), comprising only the four RobotReviewer-assessable bias types: allocation concealment, blinding of outcome, blinding of participants, and random sequence generation. Only two judgment categories (low, and high/unclear) per RobotReviewer's specification. Metric is Macro-F1. The error bar represents the 95\% confidence interval.}}
\label{tab:robotreviewer}
\end{table}
This study compares previous ML models with LLMs on four specific biases, the only ones assessable by RobotReviewer \cite{robotreviewer_acl} (an SVM/CNN ensemble) due to its inaccessible training code. We also trained logistic regression and SVM models on our Cochrane Train Set (Main) \cite{robin}. Table \ref{tab:robotreviewer} shows inconclusive results : RobotReviewer, SVM, and LR perform marginally worse, if at all, than GPT-4o and Sonnet-3.5. 

The four biases RobotReviewer assesses are considered straightforward because superficial keywords often directly indicate bias risk (e.g., "opaque envelope" for low allocation concealment bias; "randomness" or "random number generator" for low random sequence generation bias), suggesting significant keyword reliance. However, biases like selective reporting or deviation bias may necessitate a more holistic consideration of the entire biomedical paper.

Traditional ML models by definition are hard to generalize to new bias names and definitions. RoB1/RoB2 guidelines list hundreds of bias names, some with definitions re-interpreted in different reviews. we treat these as paper-specific biases, manually incorporating the paper's re-interpreted definition during LLM evaluation on RoBBR. Thus, RoBBR primarily evaluates models in zero-shot settings or fine-tunes models like Llama3-8B for generalization to unseen bias definitions via semantic understanding. Furthermore, most bert-based models' 512-token context limit is ill-suited for RoBBR's biomedical studies (average 8k tokens), unlike modern LLMs with longer context windows.

\section{Conclusion}
We present RoBBR, a benchmark for measuring models' ability to assess risk-of-bias in biomedical studies. RoBBR includes two novel subtasks that evaluate a model's retrieval and reasoning abilities. The support sentence retrieval subtask is created by our fully automatic and human-validated annotation pipeline for aligning support judgments to sentences in biomedical studies. Our analysis reveals the importance of retrieval and reasoning abilities and demonstrate their impact on the a model's ability to assess risk-of-bias.

\section*{Limitations}
\label{sec:limitation}
While in this work all systematic reviews and biomedical studies are open-sourced (Creative Commons License or Public Domain), many other systematic reviews have licenses which restrict distributing their contents. Therefore, we share the entire codebase for creating the RoBBR benchmark and encourage researchers with access to non-open-source contents to create other versions of RoBBR.

In our evaluation of LLMs, while we have developed our prompts using RoBBR's train sets and investigated popular techniques such as chain-of-though reasoning etc, we acknowledge the possibility that better prompting techniques could lead to slight model improvements on RoBBR. 

\section*{Ethics Statement}
All data in RoBBR come from open-sourced systematic reviews and biomedical studies covering the domain of evidence-based biomedicine. The authors make sure no personal information is included in RoBBR by manual inspection of all datapoints. 

The RoBBR benchmark, which promotes the automation of risk-of-bias determination, might introduce over-reliance on commercial AI systems to evaluate a biomedical study's risk-of-bias.

The benchmark only includes systematic reviews and biomedical studies in English. Systems trained and evaluated on RoBBR might disadvantage the risk-of-bias determinations for non-English biomedical studies.

\bibliography{custom}

\clearpage

\cleardoublepage
\appendix
\cleardoublepage
\appendix

\section{Dataset}
\label{sec:appendix_dataset}

\subsection{Dataset License and Code License}
\label{ssec:appendix_dataset license}
The RoBBR dataset is made available under CC-BY-NC. 
A copy of the full license can be found at \href{https://github.com/RoBBR-Benchmark/RoBBR/blob/main/LICENSE.md}{RoBBR License}.

The code used in this paper is released under the MIT License. The MIT License is a permissive open-source license that allows for the free use, modification, and distribution of the code, as long as the original license is included with any derivative work. A copy of the full license can be found at \href{https://github.com/RoBBR-Benchmark/RoBBR/blob/main/LICENSE.md}{RoBBR License}.

\subsection{Dataset Hosting, Accessibility and Maintenance}
\label{ssec:appendix_dataset hosting, accessibility and maintenance}

The RoBBR dataset with its meta-data is released and can be accessed freely at \href{https://github.com/RoBBR-Benchmark/RoBBR/blob/main/LICENSE.md}{RoBBR License}. We commit to regularly maintain the dataset and codebase by incorporating user feedback. We will potentially introduce more features as part of future work in the next version of RoBBR. We confirm that the current version of RoBBR will always remain accessible at the same link. 

\subsection{Dataset Statistics}
\label{ssec:appendix_dataset statistics}

See Tables \ref{tab: task2_testset_statistics}, \ref{tab:testset_category_statistics}, \ref{tab:task4_testset_statistics} for detailed statistics of the RoBBR tasks.

\begin{table}[ht]
\centering
\setlength{\tabcolsep}{5pt}
\renewcommand{\arraystretch}{1.1}
\begin{tabular}{@{}lrrr@{}}
\toprule
\textbf{Statistic} & \textbf{min} & \textbf{avg} & \textbf{max} \\
\midrule
Tokens per report       & 2,908   & 9,215   & 19,080 \\
Sentences               & 87      & 218.6   & 400    \\
Covered Aspects         & 1       & 1.9     & 7      \\
Optimal \# of sentences & 1       & 1.4     & 5      \\
\bottomrule
\end{tabular}
\caption{{ Statistics for the Cochrane Test Set SSR (313 data points).}}
\label{tab: task2_testset_statistics}
\end{table}

\begin{table}[ht]
\centering
\footnotesize
\setlength{\tabcolsep}{4pt} 
\begin{tabular}{@{}lrrr@{}}
\toprule
\textbf{Bias Type} & \multicolumn{1}{c}{\shortstack{\textbf{Cochrane} \\ {\textbf{Test SSR}}}} 
                   & \multicolumn{1}{c}{\shortstack{\textbf{Cochrane} \\ {\textbf{Test SJS}}}}
                   & \multicolumn{1}{c}{\shortstack{\textbf{Cochr. + Non-Cochr.} \\ {\textbf{Test Main}}}} \\
\midrule
Selection   & 157 & 130 &  933 \\
Attrition   &  46 &  98 &  629 \\
Performance &  54 &  87 &  309 \\
Detection   &  61 &  82 &  645 \\
Reporting   &  14 &  80 &  594 \\
Deviation   &   0 &   0 &  331 \\
\bottomrule
\end{tabular}
\caption{{ Test set category statistics}}
\label{tab:testset_category_statistics}
\end{table}

\begin{table}[htbp]
    \centering
    \renewcommand{\arraystretch}{1.15}  % extra vertical spacing
    \setlength{\tabcolsep}{4pt}
    \begin{tabular}{l r r r}
\toprule
\textbf{Statistic} &
\shortstack{\textbf{Cochrane}\\\textbf{Train}} &
\shortstack{\textbf{Non\hspace{-0.5pt}-Cochr.}\\\textbf{Test}} &
\shortstack{\textbf{Cochrane}\\\textbf{Test}} \\
\midrule
Points & 774 & 2 489 & 906 \\
\midrule
\multicolumn{4}{l}{Tokens per report:}\\
\quad min & 1,991 & 1,488 & 2,274 \\
\quad avg & 9,907 & 7,791 & 8,946 \\
\quad max & 18,894 & 26,154 & 19,080 \\
\midrule
\multicolumn{4}{l}{Label distribution:}\\
\quad Low & 387 & 1,465 & 562 \\
\quad Unclear & 275 & 574 & 195 \\
\quad High & 112 & 450 & 149 \\
\bottomrule
\end{tabular}
\caption{Test-set statistics for \textbf{Risk-of-Bias Determination}. Unclear/Some concerns for Non-Cochrane test}
\label{tab:task4_testset_statistics}
\end{table}

\subsection{Dataset Collection and Processing}
\label{ssec:appendix_dataset collection and processing}
We use BioC \cite{bioc} API to download meta-reviews and papers in PMC database. We manually download the rest of the papers. We use GROBID \cite{GROBID} to parse papers from PDF to XML format. We use Stanza \cite{stanza} to split paragraphs (including table and figure captions) into sentences.

\subsection{RoBBR Structure}
\label{ssec:appendix_robbr structure}

\begin{itemize}

\item SSR\_test.json and SSR\_dev.json: Test and dev set of Support Sentence Retrieval

\begin{itemize}
    \item paper\_doi: The DOI of the paper.
    \item bias: The bias to be considered.
    \item PICO: PICO of a study in the paper, including Methods, Participants, Intervention, Outcome, and Notes.
    \item objective: The meta-analysis objective.
    \item paper\_as\_candidate\_pool: A tuple of text elements from the paper. Each text element is a sentence, a section title, a table, or a figure caption.
    \item aspects: A dictionary that maps aspect id to bias aspect.
    \item aspect2sentence\_indices: a mapping (i.e. dictionary) between aspect id and all sentence indices that independently are source of information for that aspect, as annotated by our pipeline. 
    \item sentence\_index2aspects: a mapping (i.e. dictionary) between sentence index and all aspect ids that this sentence is the source of information of.
    \item bias\_retrieval\_at\_optimal\_evaluation: This is a dictionary that contains the necessary information for evaluating the model's performance on the task Support Sentence Retrieval @ Optimal.
    \begin{itemize}
        \item optimal: A positive integer, which is the smallest number of sentences needed to cover the largest number of aspects. 
        \item one\_selection\_of\_sentences: a list of sentence indices. The list size is the optimal number. The list of sentences cover the largest number of aspects. Note, there are potentially other lists of sentences that has the same size and also cover the largest number of aspects.
        \item covered\_aspects: the list of aspects that are covered. In this case, the list of aspects covered is list of all aspects. 
    \end{itemize}
\end{itemize}

\item SJS\_test.json and SJS\_dev.json: Test and dev set of Support Judgment Selection

\begin{itemize}
    \item paper\_doi: The DOI of the paper.
    \item bias: The bias to be considered.
    \item PICO: PICO of a study in the paper, including Methods, Participants, Intervention, Outcome, and Notes.
    \item objective: The meta-analysis objective.
    \item full\_paper: The full paper content.
    \item options: The seven options for the multiple choice.
    \item label: The index of the correct option.
\end{itemize}

\item Main\_task\_test.json and Main\_task\_dev.json: Test and dev set of the main task

\begin{itemize}
    \item paper\_doi: The DOI of the paper.
    \item bias: The bias to be considered.
    \item PICO: PICO of a study in the paper, including Methods, Participants, Intervention, Outcome, and Notes.
    \item objective: The meta-analysis objective.
    \item full\_paper: The full paper content.
    \item label: One of [low,high,unclear], representing the risk level of the bias.
\end{itemize}

\end{itemize}

\section{Annotation Guideline}
\label{sec:appendix_annotation guideline}
Below, we show the annotation guideline for aspect mapping of 50 randomly sampled (aspect, full paper) pair. The four annotators form two teams of two persons, all seeing the same annotation guideline.

You have a total of 50 annotation task packets. Each task packet is a docx. file that contains the following information.

\begin{itemize}
    \item An Aspect (one piece of important information/detail).
    \item The support judgment and the bias.
    \item The doi of the paper.
    \item An indexed list of text elements of the paper (a text element could be a sentence, tables, a figure caption, etc.)
\end{itemize}

You have to pledge the following conditions are met during annotation for each task packet.
\begin{itemize}
    \item For words you are not familiar with and believe are important for comprehension, conduct the search to understand its meaning.
    \item For every text element in the list, you have to look at it and read it at least once.
    \item You cannot talk to the other annotator team about anything related to your task, including progress and insights.
    \item You should independently do the task first, and then consult with the other person in your team after completing the task.
    \item You have to take a mandatory 5 minute break after every 1 hour of performing annotation.
    \item You cannot exceed 8 hours of annotation per day.
\end{itemize}

Below is the recommended procedure for annotating each packet.
\begin{itemize}

    \item Read and understand the aspect and the support judgment first. You need to understand the context of the aspect, i.e., the role of the aspect in the support judgment given the bias. You also need to understand why this aspect is important for judging the bias. You can consult with LLM to understand this aspect.

    \item Decide what details are important in the aspect. Geo, temporal, and numerical data are all important details.

    \item For each text element, you need to decide if a significant amount of important details can be implied from the text element.

    \begin{itemize}
        \item When deciding the level of implication, consider how the text element can help judge the bias. You should not make complicated implications, i.e., can you see the aspect from the text element within 30 seconds? If not, then it is not a match.
    
        \item Pay attention to acronyms, abbreviations, or different presentation formats of the same information.
    
        \item Even if you find significant information that can be implied from the text element, you have to make sure the text element is in the same context as the aspect.

        \begin{itemize}
            \item Same context typically refers to the same study or experiment, and for numerical results in Table, it means row and column must indicate the same setting.
    
            \item Check if the text element refers to the same experiment as the aspect. Since different experiments could be in one paper. Maybe the text element does not refer to any experiment at all.
        \end{itemize}
        
        \item If you suspect that there might be a relationship between the text element and the aspect but do not understand the meaning of the text element, you can use LLM for help. However, you must justify the response from LLM, and you should not rely on the response from LLM.
    \end{itemize}

    \item If a text element is a Table, refer to the actual Table in the pdf for better understanding. However, only consider the information in the text element.

    \item After you have independently finished all 50 tasks, you should talk to the other person in your team following these procedures:

    \begin{itemize}
        \item Go through task 0-49.

        \item Resolve your difference, check if you made a mistake, or if you missed something. If you made a conceptual error (e.g., fail to understand some terminology), you may have to quickly go through the paper again.

        \item For sentences that you cannot resolve your difference after discussion, i.e., one person says yes and the other person says no, or if both people are unsure, you should include them in your final list of decisions.
        
    \end{itemize}

    \item Ultimately, you and your teammate should collaboratively arrive at a consensus for each of the 50 tasks. Write the collective answers in the answer file provided.

End of Annotation Guideline.

\subsection{Annotator Training}
\label{ssec:appendix Annotator Training}
The annotation teams each comprised two graduate students. To ensure a thorough understanding of bias names and risk levels, all annotators were trained and briefed on their definitions prior to beginning annotation. A key part of their preparation involved an in-depth reading of five to ten full-length meta-analyses from Cochrane Train (Main), including their extensive supplementary materials (often over 100 pages), to gain familiarity with relevant terminology and details. Although our annotation guidelines allowed for the use of Large Language Models (LLMs) to facilitate understanding of unfamiliar subject matter, annotators infrequently consulted these tools. Their preference was to use search engines (with AI overview features disabled). On the rare occasions LLMs were employed, their function was strictly confined to assisting with the comprehension of unfamiliar technical details. Annotators were monitored and were expressly forbidden from soliciting LLM decisions or opinions regarding annotation choices. It is important to note that LLMs were infrequently utilized and only as a supplementary tool for assisting the understanding of unfamiliar subject matters.

\end{itemize}

\section{Task Descriptions and Visualizations}
\label{sec:appendix task visual}
We provide visual illustrations of the three tasks.

\subsection{Main Task: Risk-of-Bias Determination}
The model is given the whole biomedical paper, study characteristics (i.e. participants, intervention methods, comparators, outcome, etc.), the objective/topic of the systematic review, one bias name and definition, and general guideline that explains how to classify the risk level as high, low or unclear/some concern, and is asked to determine the risk level of bias. Note, the definition is based on the specific bias name and risk level. Our categorization of different bias names into six main categories does not affect the definition of individual biases. See Table \ref{tab:bias-categorization} for examples of bias categorization.

\begin{table*}[!t]
    \centering
    \footnotesize
    \begin{tabularx}{\textwidth}{@{}>{\raggedright\arraybackslash}X
                                     >{\raggedright\arraybackslash}X@{}}
    \toprule
    \textbf{Bias Name} & \textbf{Bias Category} \\
    \midrule
    random sequence generation (selection bias) & selection bias \\
    bias arising from the randomization process & selection bias \\
    bias due to deviations from intended intervention & deviation bias \\
    bias due to missing outcome data & attrition bias \\
    incomplete outcome data (attrition bias) all outcomes & attrition bias \\
    similarity of other baseline characteristics (selection and performance bias) & selection bias / performance bias \\
    bias in measurement of the outcome & detection bias \\
    blinding of outcome assessment (detection bias) all outcomes & detection bias \\
    blinding for adverse events (performance and detection bias) & performance bias / detection bias \\
    blinding of participants and personnel (performance bias) & performance bias \\
    selective reporting (reporting bias) & reporting bias \\
    bias in selection of the reported result & reporting bias \\
    blinding (performance bias and detection bias) all outcomes & performance bias / detection bias \\
    \bottomrule
    \end{tabularx}
    \caption{Examples of Bias Categorization}
    \label{tab:bias-categorization}
\end{table*}

\subsection{Subtask: Support Sentence Retrieval}
\begin{figure*}[ht]
    \centering
    \includegraphics[width=\linewidth]{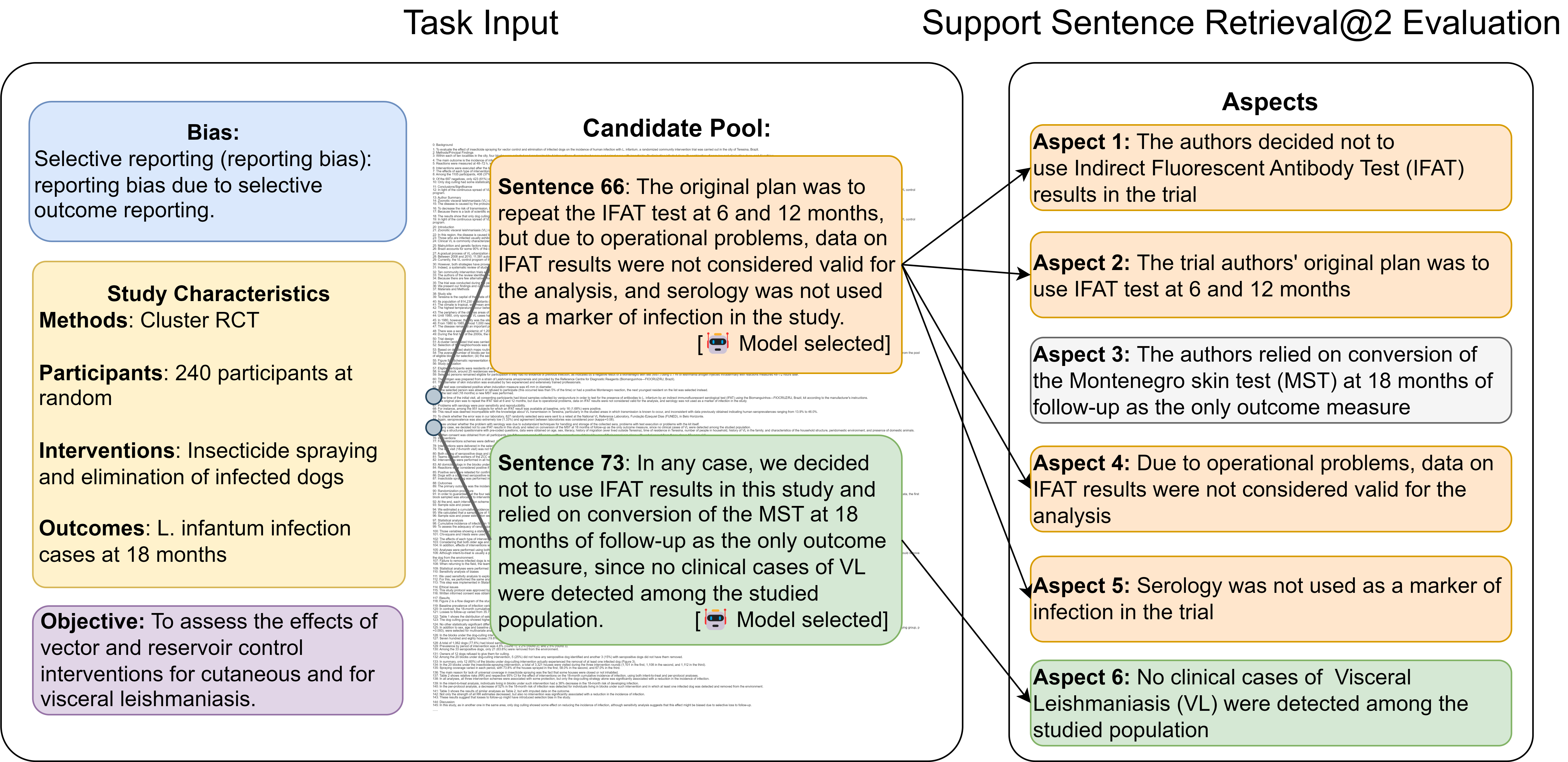}
    \caption{Subtask: Support Sentence Retrieval. In this task, the goal is to retrieve no more than the optimal number of sentences from the biomedical report which support a risk of bias judgment. Performance is evaluated against a support judgment, which has been split into aspects.}
    \label{fig:task2}
\end{figure*}

When judging the risk of bias for a specific study, the review authors provide support judgments which justify their risk of bias rating. These support judgments are grounded in specific aspects of the study, which the support judgment describes. This  aims to test a model's ability to retrieve the correct source for the support judgment from a paper's text. Note the model is instructed to retrieve no more than the optimal number of sentences. The optimal number can be calculated from Aspect Sentence Mapping and is already included in our dataset. See Figure \ref{fig:task2} for an illustration.

\subsection{Subtask: Support Judgment Selection}

\begin{figure*}[ht]
    \centering
    \includegraphics[width=\linewidth]{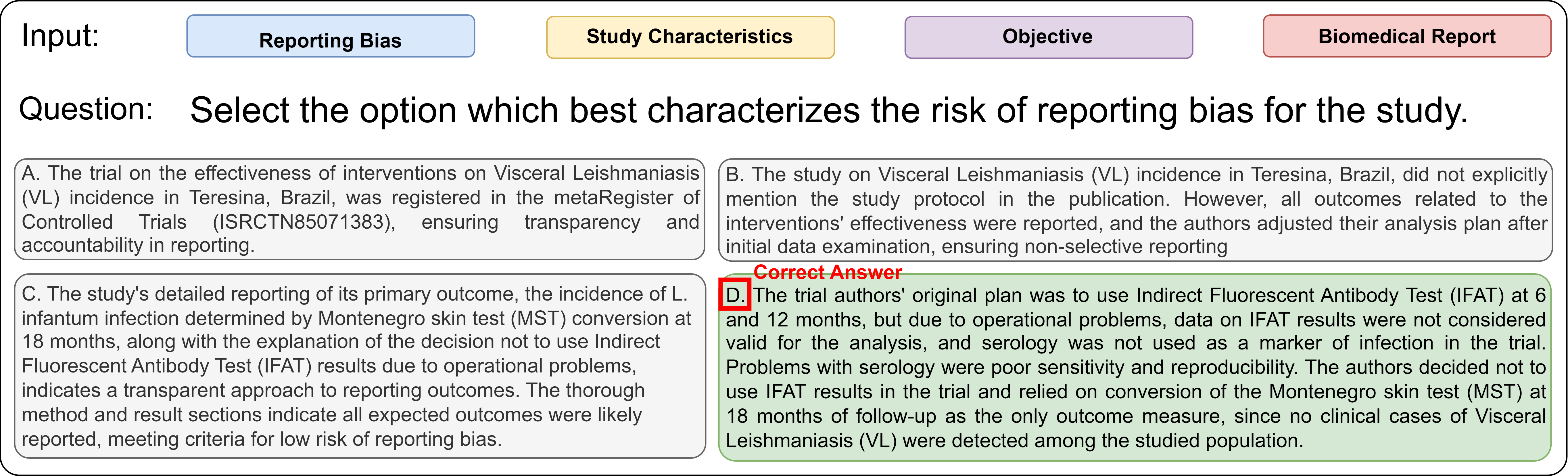}
    \caption{Subtask: Support Judgment Selection. In this task, the model is shown a paper, and asked which of 7 support judgments provides the best explanation of the paper's risk of bias. Only four options are shown here for illustration purposes. Both options C and D refer to the same information from the paper, in this case the operational challenges affecting the trial's methodology, but describe different reasoning given this information. This demonstrates that retrieving the correct information from the paper is not sufficient for solving this task.}
    \label{fig:task3}
\end{figure*}

Figure \ref{fig:task3} shows an example of a multiple-choice question with one correct answer and three synthetically generated answers. Both options C and D refer to the same information from the paper, in this case the operational challenges affecting the trial's methodology, but describe different reasoning given this information. This demonstrates that retrieving the correct information from the paper is not sufficient for solving this task.

\cleardoublepage

\cleardoublepage

\section{Support Sentence Retrieval Prompt, Optimization and Example}
\label{sec:appendix_task 2 prompt optimization}

\subsection{Aspect Decomposition Prompt}
\label{sec:appendix_aspect decomposition prompt}
\begin{figure*}[htbp]
    \centering
    \includegraphics[width=\linewidth]{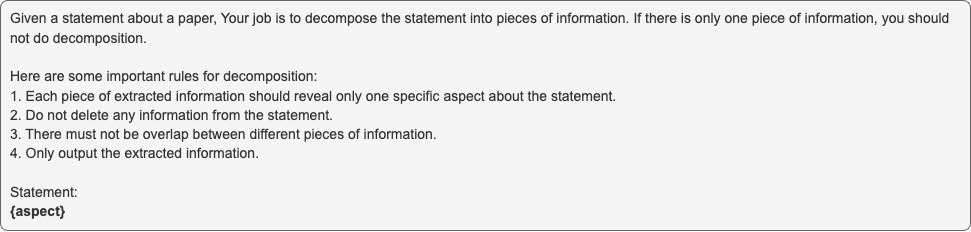}
    \caption{{\small Prompt for decomposing support judgment into aspects.}}
    \label{fig:aspect_decomposition_prompt}
\end{figure*}

See Figure \ref{fig:aspect_decomposition_prompt} for the aspect decomposition prompt.

\subsection{Aspect Filtering Prompt}
\label{sec:appendix_aspect filtering prompt}
See Figure \ref{fig:filter_aspect} for prompt to filter non-specific commentaries.

\subsection{Aspect-to-Sentence Mapping Prompt and Optimization}
\label{sec:appendix_aspect-to-sentence mapping prompt and optimization}

\begin{figure*}[htbp]
    \centering
    \includegraphics[width=\linewidth]{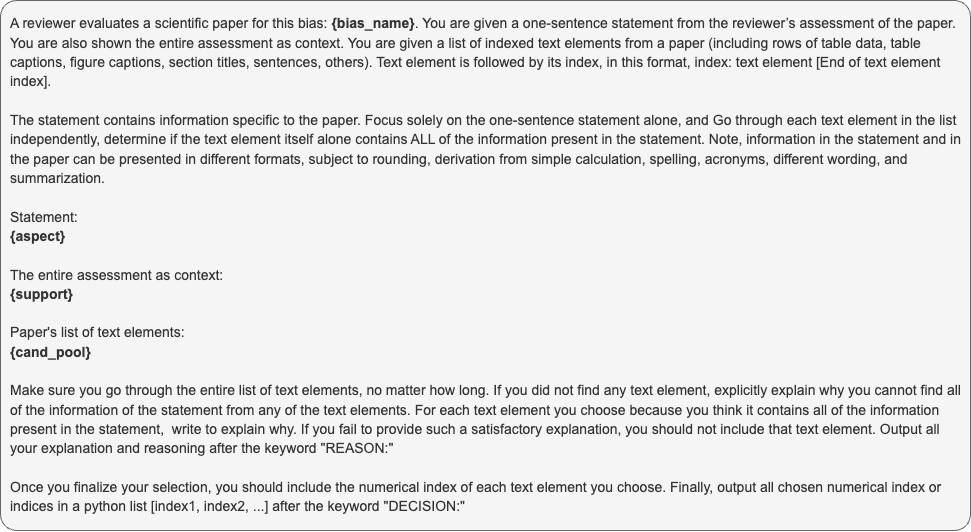}
    \caption{{\small Prompt for generating synthetic options.}}
    \label{fig:aspect_tagging_prompt}
\end{figure*}

See Figure \ref{fig:aspect_tagging_prompt} for the prompt optimized on the development set.

To enhance the agreement between human annotators and the GPT-4-0125-preview model, we developed and tested various prompt engineering strategies on a development set consisting of 30 unique (aspect, paper) pairs. It is important to note that no paper or aspect from the development set overlapped with the 50 tasks in the hypothesis testing set.

GPT-4-0125-preview was chosen due to its robust instruction-following capabilities and our familiarity with its performance across different settings. Our experimentation revealed that using in-context learning examples did not improve agreement rates and tended to cause overfitting. Instead, we found that embedding the same instruction at both the beginning and the end of the prompt effectively helped the model maintain focus on the task of identifying text elements that significantly cover the details specified in an aspect.

We also implemented chain-of-thought reasoning, prompting GPT-4 to articulate its thought process following a specific keyword. This approach not only enhanced the quality of the model’s reasoning but also stabilized its performance and reduced common-sense errors.

To address the issue of entity forgetting in long-context tasks (a typical paper might contain around 5000 tokens), we employed a sliding window technique. Each window, containing 10 text elements with a 5-element overlap, allowed GPT-4 to process and evaluate each text element within a manageable context size. The hyperparameters, window length and overlap size, were optimized using the development set. This overlapping approach ensures that each text element is evaluated twice, significantly reducing the likelihood of false negatives.

Figure \ref{fig:aspect_tagging_prompt} illustrates the final prompt template used to match text elements (such as sentences, tables, figure captions, etc.) with the aspects. Each template inputs one aspect, the 10 text elements within a sliding window, and the contextual background.

This methodology not only ensures high fidelity in aspect-text alignment but also leverages the model’s capabilities to provide consistent and accurate annotations across extensive text bodies.

\subsection{A Motivating Example}
\label{sec:appendix_a motivating example}

Here we provide an example of how we build the support sentence retrieval task.\\
\\
\textbf{Bias}\\
Selective reporting (reporting bias)\\
\\
\textbf{Support Judgment For the Bias}\\
The trial authors' original plan was to use Indirect Fluorescent Antibody Test (IFAT) at 6 and 12 months, but due to operational problems, data on IFAT results were not considered valid for the analysis, and serology was not used as a marker of infection in the trial. Problems with serology were poor sensitivity and reproducibility. The authors decided not to use IFAT results in the trial and relied on conversion of the Montenegro skin test (MST) at 18 months of follow-up as the only outcome measure, since no clinical cases of Visceral Leishmaniasis (VL) were detected among the studied population.\\
\\
\textbf{Decomposition of Support Judgment into Aspects}

\begin{itemize}[nosep]
    \item Aspect 1: The trial authors' original plan was to use IFAT test at 6 and 12 months
    \item Aspect 2: Due to operational problems, data on IFAT results were not considered valid for the analysis
    \item Aspect 3: Serology was not used as a marker of infection in the trial
    \item Aspect 4: Problems with serology were poor sensitivity and reproducibility.
    \item Aspect 5: The authors decided not to use Indirect Fluorescent Antibody Test (IFAT) results in the trial
    \item Aspect 6: The authors relied on conversion of the Montenegro skin test (MST) at 18 months of follow-up as the only outcome measure 
    \item Aspect 7: No clinical cases of VL were detected among the studied population
\end{itemize}

\begin{figure*}[htbp]
    \centering
    \includegraphics[width=\linewidth]{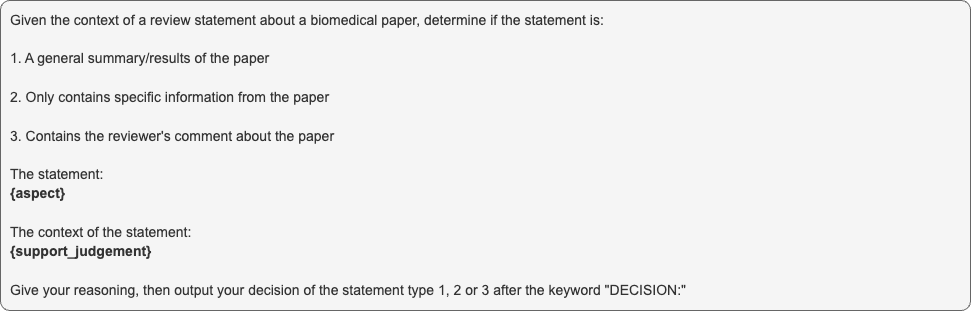}
    \caption{{\small Prompt for filtering non-specific commentaries.}}
    \label{fig:filter_aspect}
\end{figure*}

\noindent \textbf{Aspect Filtering}\\
Using prompt \ref{fig:filter_aspect}, Aspect 4 is filtered since it is a commentary of the reviewer.\\
\\
\textbf{Mapping Aspects to Sentences in Report}\\
Utilizing the procedure described in section \ref{sec:appendix_aspect-to-sentence mapping prompt and optimization}, we map the remaining 6 aspects to all sentences in the report \cite{example_paper}.\\
\\
We only show sentences from the report that are matched with at least one aspect.\\
\begin{itemize}[nosep]
    \item Sentence 4: The main outcome is the incidence of infection assessed by the conversion of the Montenegro skin test (MST) after 18 months of follow-up in residents aged $\geq$1 year with no previous history of visceral leishmaniasis (VL). (\textbf{Mapped with aspect 6})
    \item Sentence 66: The original plan was to repeat the IFAT test at 6 and 12 months, but due to operational problems, data on IFAT results were not considered valid for the analysis, and serology was not used as a marker of infection in the study. (\textbf{Mapped with aspect 2, 3, 5, 7})
    \item Sentence 73: In any case, we decided not to use IFAT results in this study and relied on conversion of the MST at 18 months of follow-up as the only outcome measure, since no clinical cases of VL were detected among the studied population. (\textbf{Mapped with aspect 1, 3, 5, 6})
\end{itemize}

\noindent \textbf{Aspect Recall Ratio @ Optimal}\\
One of our evaluation metric, Aspect Recall Ratio @ Optimal, measures the  the optimal number of sentences to cover all aspects. In this example, we only need two sentences, sentence 66 and sentence 73, to cover all aspects.

\cleardoublepage

\section{Task-specific Fine-tuning}
\label{appendix:fine-tuning}

We fine-tune Llama-3-8B on the three train sets of RoBBR's three tasks using LoRA \cite{lora}, on 1 node of 8 NVIDIA H-100. Our fine-tuning hyperparameters are standard: LoRA rank $=8$, LoRA alpha $=16$, a batch size $=16$, with the AdamW optimizer (weight decay $=0.01$), and a cosine learning rate scheduler with 10\% warmup. Llama-3-8B is fine-tuned for 1 epoch using cross entropy loss. The model is asked to predict the label (i.e., risk-level for main task, sentence indices for SSR, correct support judgment for SJS). We compared different fine-tuning techniques, and found when we train model to predict both the label and a paragraph of reasoning text, the optimal training loss is achieved on the train sets, since it increases the amount of training tokens directly go through cross entropy loss \cite{distill_step_by_step}. We use support judgment as proxy for reasoning text.

\section{Analysis: Two Types of Long-Context Problems}
\label{sec: appendix long context local vs global}
\begin{figure}[htbp]
    \centering
    \includegraphics[width=\linewidth]{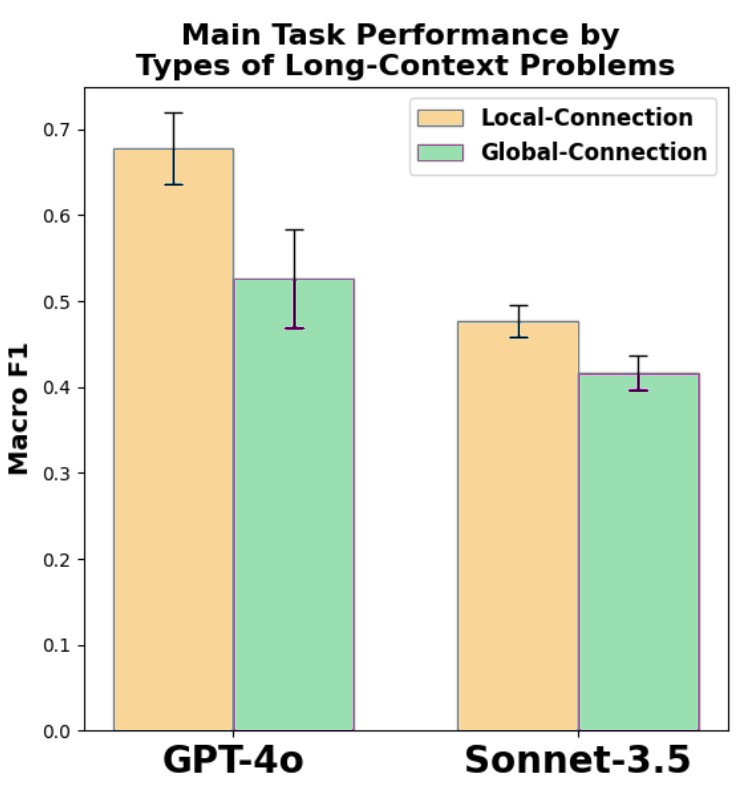}
    \caption{{\small Local-connection problems are easier for LLMs.}}
    \label{fig:long_context}
\end{figure}

An active research direction in long-context modeling is to categorize different types of long-context problems \cite{rag_2024}. Using our annotation pipeline's Aspect-to-Sentence mapping, we can naturally categorize RoBBR's datapoints into two types by the optimal number of support sentences used to cover all aspects. 

\noindent\textbf{Local-Connection Problem:}\\
If one support sentence can cover all aspects that form the support judgment, it means this support sentence is all a model needs to reach the same support judgment. While the main task for this datapoint is still a long-context problem because RoBBR's average document length exceeds 8k tokens, it can be solved by locating one specific sentence from the report. There are 213 local-connection datapoints in RoBBR. 

\noindent\textbf{Global-Connection Problem:}\\
If the support judgment cannot be traced back to one single sentence, some degree of global understanding of the report is required. There are 693 global-connection datapoints in RoBBR.

In Figure \ref{fig:long_context}, we calculate the average LLM accuracy per type, and found evidence that suggests local-connection problems are easier for both GPT-4o and Sonnet-3.5. The intuition is that a problem becomes harder when a more global understanding of the entire long context is required.

\section{Error Analysis}
\label{sec:appendix_error_analysis}

To better illustrate that \textbf{GPT-4o} and \textbf{Sonnet-3.5} possess different reasoning strengths, we compare their reasoning traces when both are given the same ground-truth retrieval results.

\begin{quote}\small
\textbf{Bias:} Allocation concealment (selection bias) \\

\textbf{Bias Definition and Selection Criteria for Risk Level:}

Selection bias (biased allocation to interventions) due to inadequate concealment of allocations prior to assignment.

low risk: Participants and investigators enrolling participants could not foresee assignment because one of the following, or an equivalent method, was used to conceal allocation: Central allocation (including telephone, web-based and pharmacy-controlled randomization); Sequentially numbered drug containers of identical appearance; Sequentially numbered, opaque, sealed envelopes.

high risk: Participants or investigators enrolling participants could possibly foresee assignments and thus introduce selection bias, such as allocation based on: Using an open random allocation schedule (e.g. a list of random numbers); Assignment envelopes were used without appropriate safeguards (e.g. if envelopes were unsealed or non-opaque or not sequentially numbered); Alternation or rotation; Date of birth; Case record number; Any other explicitly unconcealed procedure.

unclear risk: Insufficient information to permit judgement of 'Low risk' or 'High risk'. This is usually the case if the method of concealment is not described or not described in sufficient detail to allow a definite judgement – for example if the use of assignment envelopes is described, but it remains unclear whether envelopes were sequentially numbered, opaque and sealed. \\

\textbf{Retrieved Sentence:}

Patients were randomly assigned to the control or experimental group following a simple randomization procedure (computerized random numbers) achieved using opaque envelopes.

Physical therapy was started under consultation with the rehabilitation medicine department after wound healing, and the allocation information to each group was not provided to reduce bias. \\

\textbf{Sonnet-3.5:}

REASON: The study describes using a computerized random number generator for randomization, which is a good start for reducing selection bias. They also mention using opaque envelopes, which aligns with one of the low-risk criteria (sequentially numbered, opaque, sealed envelopes). However, the description doesn't explicitly state that the envelopes were sequentially numbered or sealed. Additionally, while they mention that allocation information was not provided to reduce bias, it's not entirely clear if this refers to blinding after assignment or concealment before assignment.

The use of opaque envelopes suggests an attempt at concealment, but without the full details of how these envelopes were used (sequentially numbered, sealed), we can't be certain that the concealment was adequate to prevent foresight of assignments. The computerized randomization is a positive factor, but the implementation of the envelope system is crucial for proper concealment.

Given these factors, there is some evidence of attempts to conceal allocation, but not enough information to confidently classify it as low risk. At the same time, there's no clear indication of high risk practices.

DECISION: unclear \\

\textbf{GPT-4o:}

REASON: The study describes the use of opaque envelopes for allocation concealment, which is a method that can prevent participants and investigators from foreseeing assignments, provided that the envelopes are sequentially numbered, opaque, and sealed. The highlighted sentences indicate that opaque envelopes were used, but do not specify whether they were sequentially numbered and sealed. However, the mention of using opaque envelopes suggests an attempt to conceal allocation, and the additional information that allocation information was not provided to reduce bias further supports the intention to prevent foreseeing assignments. Despite the lack of explicit detail about sequential numbering and sealing, the description leans towards a method that is generally considered adequate for allocation concealment.

DECISION: low
\end{quote}

The provided example demonstrates that Sonnet-3.5 is better in reasoning compared to GPT-4o. Sonnet-3.5 correctly identifies critical information gaps in the allocation concealment process—while opaque envelopes were used, there's no confirmation they were sealed and sequentially numbered, which are important for allocation concealment according to the decision criteria. Sonnet-3.5 also recognizes that the mention of withheld allocation information could indicate post-assignment blinding rather than proper pre-assignment concealment. In contrast, GPT-4o fails to identify these crucial gaps and makes unwarranted assumptions about concealment adequacy without supporting evidence for key criteria. This example clearly illustrates Sonnet-3.5's better reasoning capabilities.

\section{SJS Synthetic Option Generation Procedure and Prompt}
\label{sec:appendix_synthetic option generation procedure and prompt}

\begin{figure*}[htbp]
    \centering
    \includegraphics[width=\linewidth]{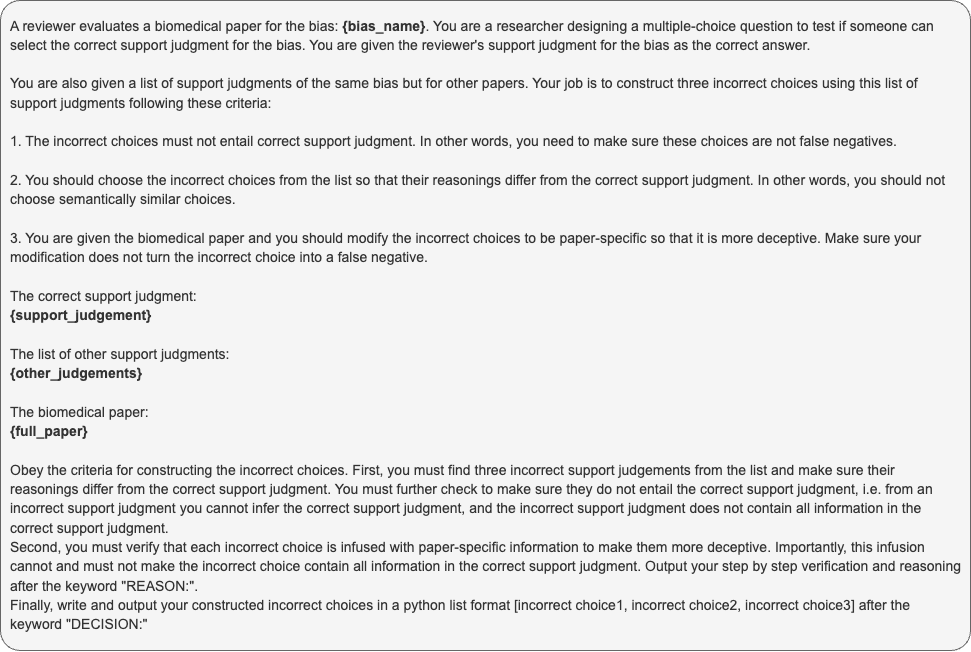}
    \caption{{\small Prompt for generating synthetic options.}}
    \label{fig:task3_synthetic_option_prompt}
\end{figure*}

\begin{figure*}[htbp]
    \centering
    \includegraphics[width=\linewidth]{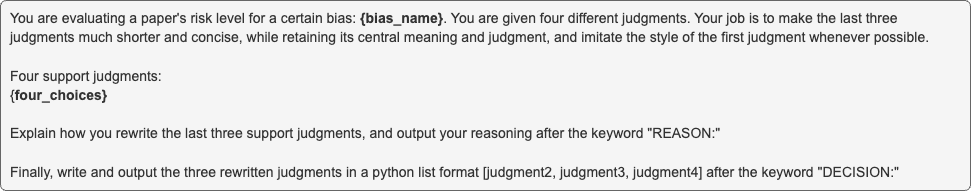}
    \caption{{\small Prompt for shortening the synthetic option.}}
    \label{fig:task3_shorten_synthetic_options_prompt}
\end{figure*}

We first generate three detailed synthetic options that are specifically crafted to imitate the support judgments from other papers concerning the same type of bias, ensuring that they are tailored to be relevant to the specific paper in question while maintaining the foundational reasoning. See Figure \ref{fig:task3_synthetic_option_prompt} for the prompt template. Following this, we condense these detailed options into shorter versions that preserve the original meaning, using prompt \ref{fig:task3_shorten_synthetic_options_prompt}. To prevent heuristic algorithms from solving the task easily, we randomly select either the long or short version of each synthetic option to include in the multiple choice questions.

\begin{figure*}[htbp]
    \centering
    \includegraphics[width=\linewidth]{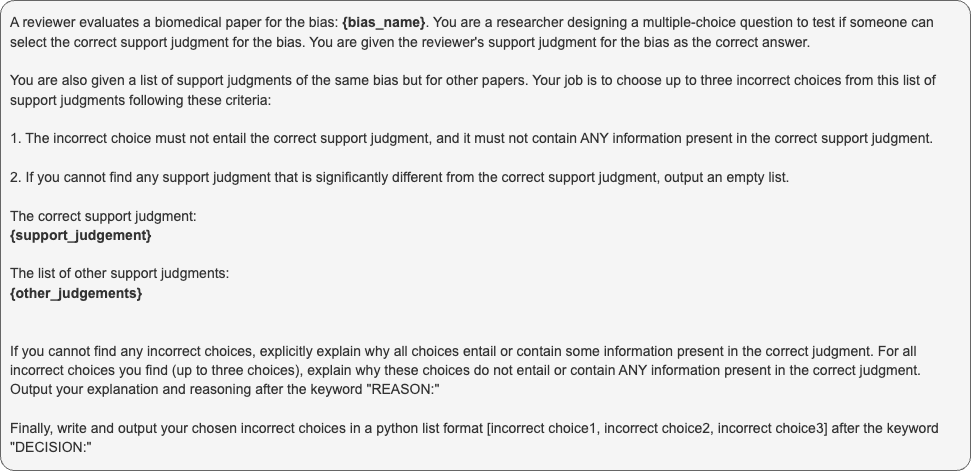}
    \caption{{\small Prompt for finding negative options from other papers’ support judgments of the same bias.}}
    \label{fig:task3_real_incorrect_prompt}
\end{figure*}

For selecting the three options derived from other papers' support judgments within the same bias category, we use prompt \ref{fig:task3_real_incorrect_prompt}.

\begin{figure*}[htbp]
    \centering
    \includegraphics[width=\linewidth]{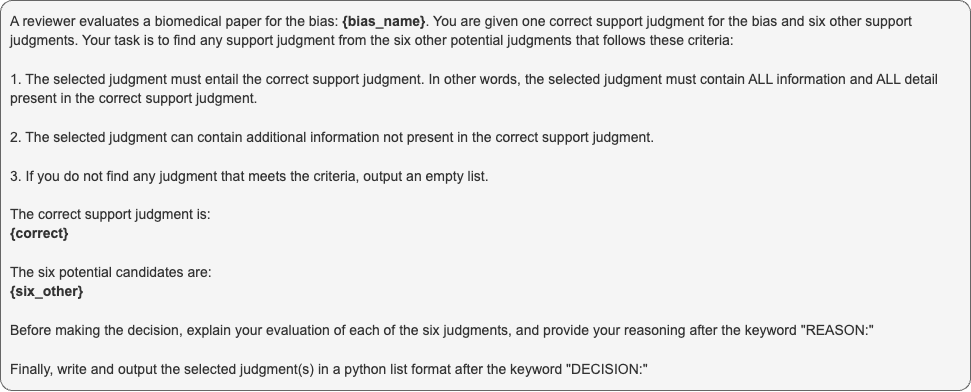}
    \caption{{\small Prompt for filtering false negatives.}}
    \label{fig:task3_filter_false_negatives_prompt}
\end{figure*}

Finally, once the six incorrect options are constructed, we conduct a manual review with the help of prompt \ref{fig:task3_filter_false_negatives_prompt} on all data points. This review ensures that the options are incorrect and not false negatives. 

\cleardoublepage

\section{Experiment Details}
\label{sec:appendix_experiment details}

\subsection{Prompt Optimization}
\label{ssec:appendix_prompt optimization}
All evaluation prompts are optimized using disjoint development set for each task. Each prompt includes specialized instructions designed to elicit chain-of-thought reasoning, thereby enhancing the models' reasoning abilities. These instructions are repeated twice: once at the beginning and once at the end of the prompt, ensuring that the models retain the instructions even after processing the entire paper. Empirical evidence shows that few-shot in-context learning does not improve model performance. Consequently, evaluation prompts do not incorporate in-context learning. It is important to note that prompt optimization is not tailored to any specific large language model; instead, the aggregated performance across all models is used to guide the optimization of evaluation prompts. 

\subsection{Prompts for Evaluation}
\label{ssec:appendix_prompts for evaluation}

\subsubsection{Prompt to Evaluate Support Sentence Retrieval (SSR)}
\label{sssec:appendix_prompt to evaluate bias retrieval}
\begin{figure*}[htbp]
    \centering
    \includegraphics[width=\linewidth]{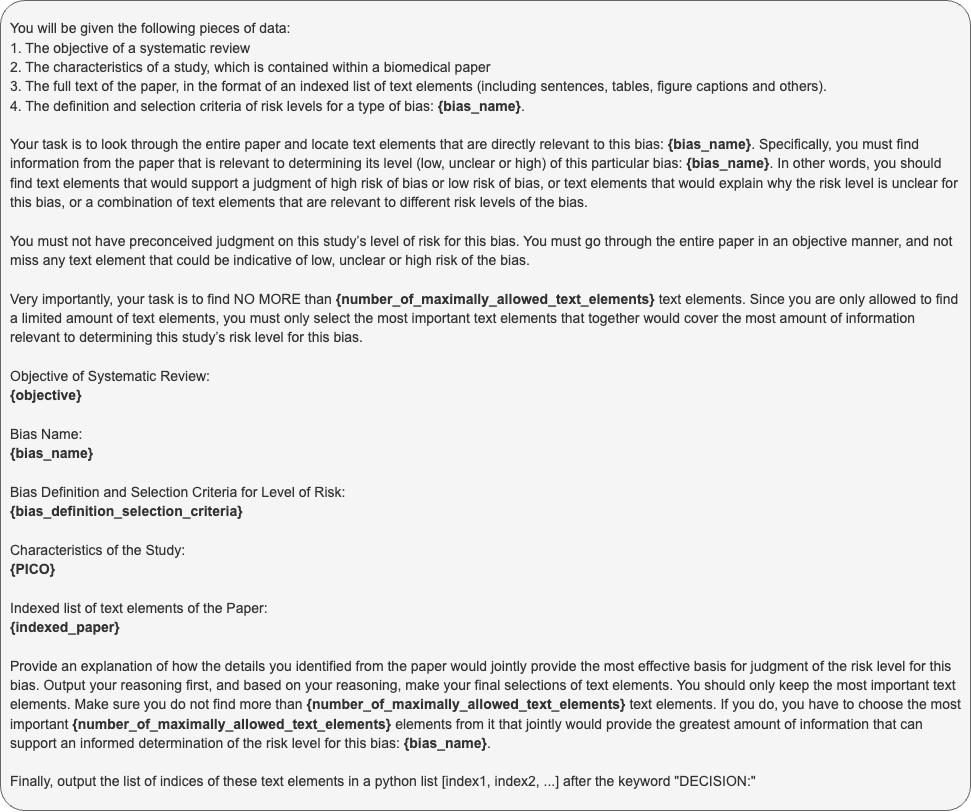}
    \caption{{\small Prompt for Support Sentence Retrieval (SSR) evaluation.}}
    \label{fig:task2_retrieval_normal_retrieval_evaluation_prompt}
\end{figure*}

\begin{figure*}[htbp]
    \centering
    \includegraphics[width=\linewidth]{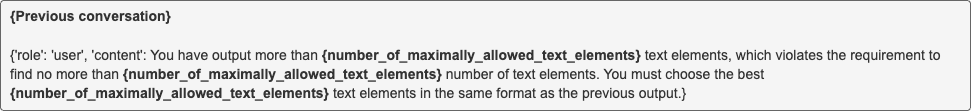}
    \caption{{\small Prompt for Support Sentence Retrieval (SSR) evaluation when the model output more than required.}}
    \label{fig:task2_retrieval_multiturn_prompt_for_exceeding_limit_regeneration_evaluation_prompt}
\end{figure*}

See Figure \ref{fig:task2_retrieval_normal_retrieval_evaluation_prompt} for prompt to evaluate Support Sentence Retrieval (SSR). We use an additional multi-turn prompt \ref{fig:task2_retrieval_multiturn_prompt_for_exceeding_limit_regeneration_evaluation_prompt} when the model outputs more than required.

\newpage
\subsubsection{Prompt to Evaluate Support Judgment Selection (SJS)}
\label{sssec:appendix_prompt to evaluate support judgment selection}
\begin{figure*}[htbp]
    {\setlength{\abovecaptionskip}{2pt} \setlength{\belowcaptionskip}{0pt} 
    \centering
    \includegraphics[width=\linewidth]{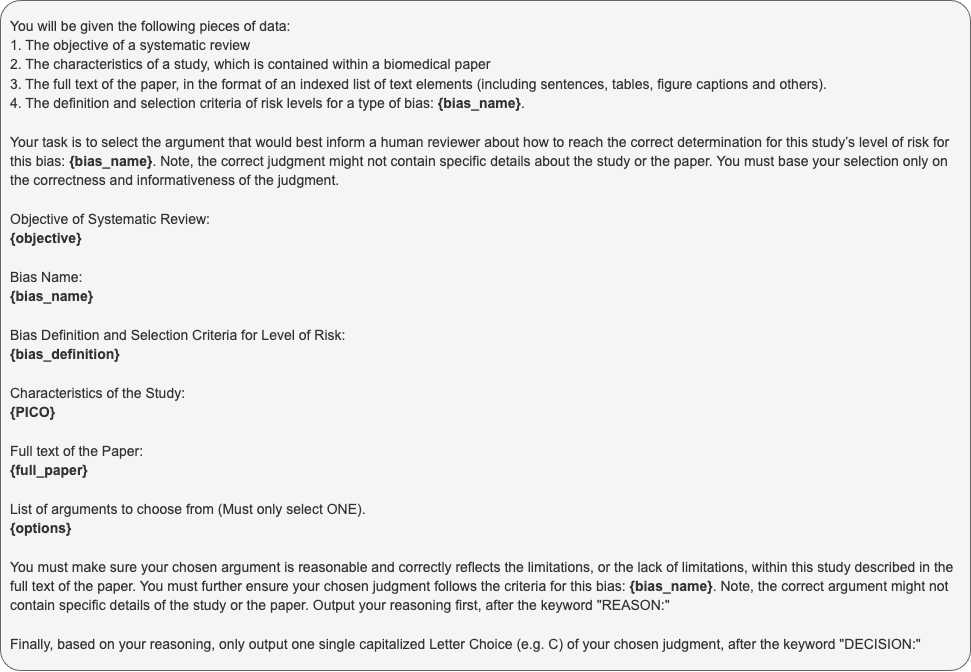}
    \caption{{\small Prompt for Support Judgment Selection (SJS) evaluation.}}
    \label{fig:task3_multiple_choice_evaluation_prompt}}
\end{figure*}

See Figure \ref{fig:task3_multiple_choice_evaluation_prompt} for prompt to evaluate Support Judgment Selection (SJS).

\subsubsection{Prompt to Evaluate Risk-of-Bias Determination (Main Task)}
\begin{figure*}[htbp]
    {\setlength{\abovecaptionskip}{2pt} \setlength{\belowcaptionskip}{0pt} 
    \centering
    \includegraphics[width=\linewidth]{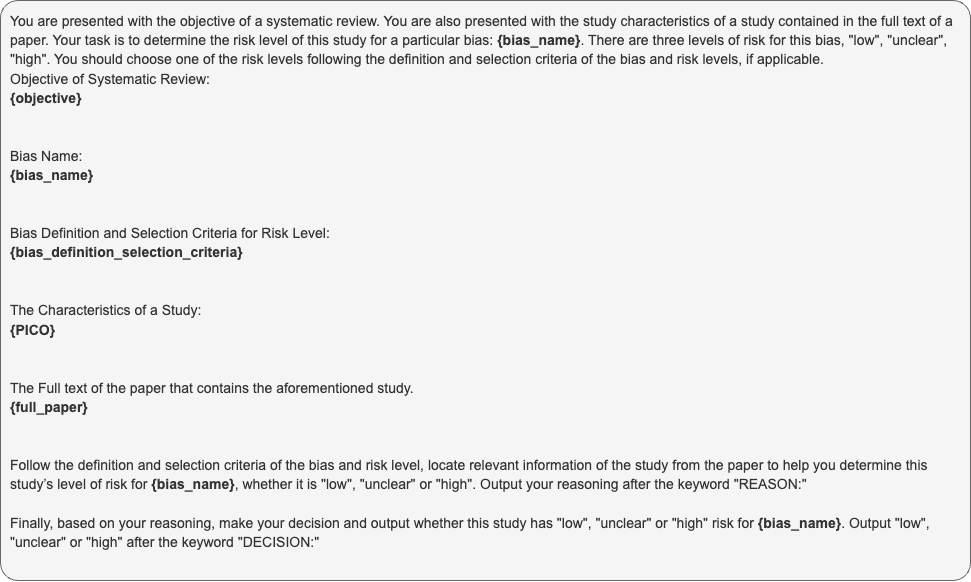}
    \caption{{\small Prompt for Risk-of-Bias Determination (Main Task) evaluation.}}
    \label{fig:task4_risk-level_determination_evaluation_prompt}}
\end{figure*}

See Figure \ref{fig:task4_risk-level_determination_evaluation_prompt} for prompt to evaluate Risk-of-Bias Determination (Main Task).
\label{sssec:appendix_prompt to evaluate risk level determination}

\subsection{Prompt for Risk-of-Bias Determination (Main Task) Given Retrieval Results}
\label{ssec:appendix_prompts for main with retrieval}

\begin{figure*}[htbp]
    {\setlength{\abovecaptionskip}{2pt} \setlength{\belowcaptionskip}{0pt} 
    \centering
    \includegraphics[width=\linewidth]{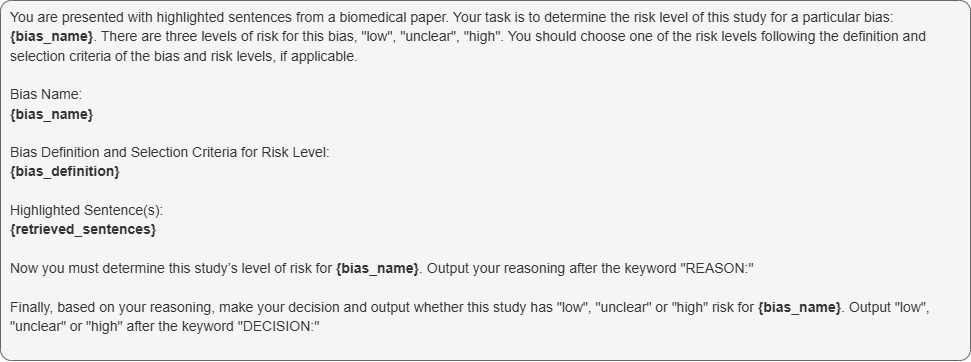}
    \caption{{\small Prompt for Risk-of-Bias Determination (Main Task) Given Retrieval Results.}}
    \label{fig:risk-level_determination_with_retrieval_prompt}}
\end{figure*}

See Figure \ref{fig:risk-level_determination_with_retrieval_prompt} for prompt for Risk-of-Bias Determination (Main Task) given retrieval results.

\subsection{Prompt for OpenAI o4-mini to identify support sentences}
\label{ssec:appendix_prompts o4 mini}

\begin{figure*}[htbp]
    {\setlength{\abovecaptionskip}{2pt} \setlength{\belowcaptionskip}{0pt} 
    \centering
    \includegraphics[width=\linewidth]{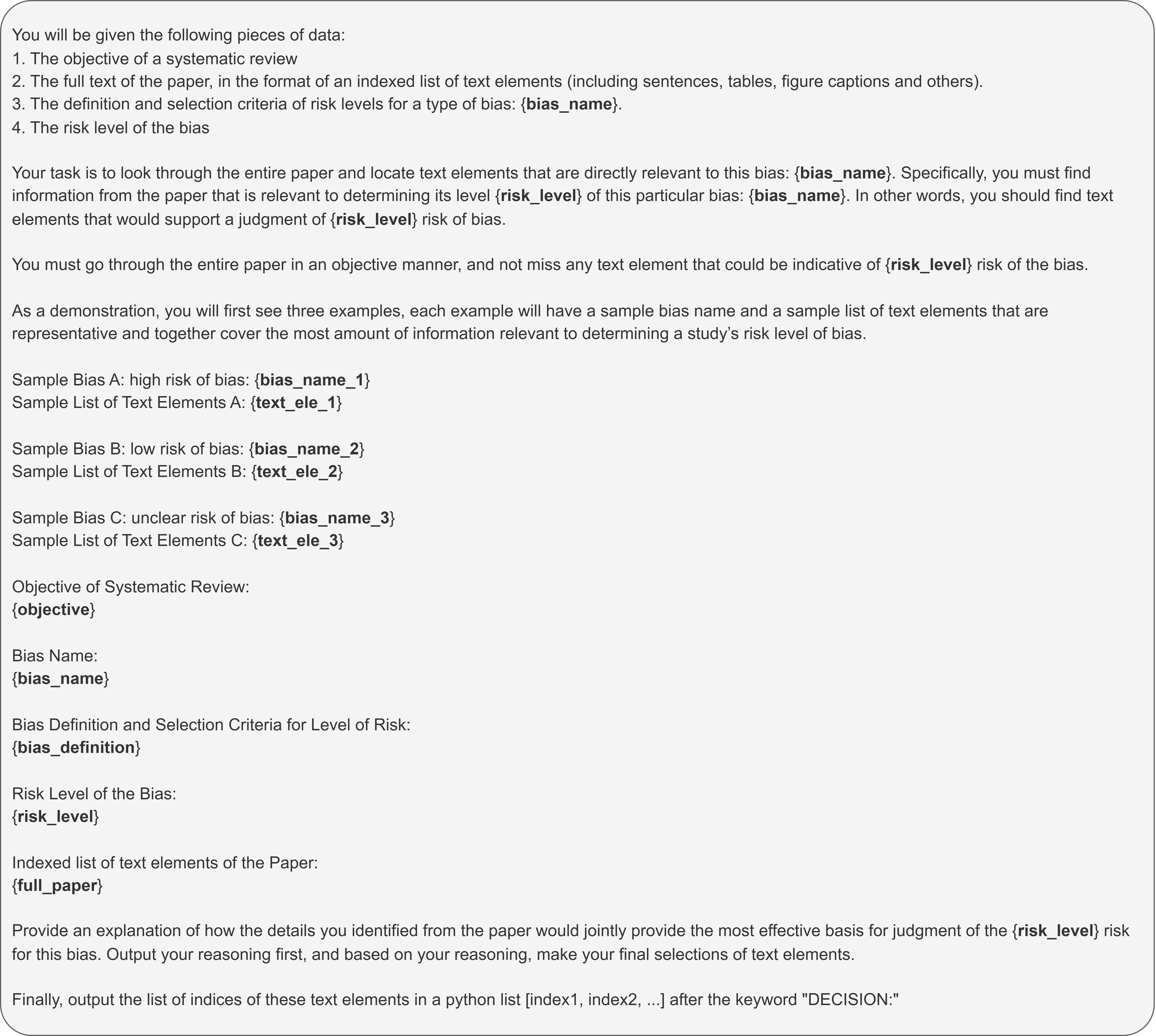}
    \caption{{\small Prompt for OpenAI o4-mini to identify support sentences.}}
    \label{fig:prompt for o4 mini findq uasi groundtruth}}
\end{figure*}

See Figure \ref{fig:prompt for o4 mini findq uasi groundtruth} for prompt for OpenAI o4-mini to identify support sentences

\subsection{Results with Standard Error}
\label{ssec:appendix_results with standard error}
We provide bootstrapped standard errors for the experimental results in Tables \ref{tab:task_2_sentence_retrieval_results_ste},\ref{tab:task_3_support_judgment_selection_results_ste},\ref{tab:task_4_risk_level_determination_results_ste}.

\begin{table*}[htbp]
\vspace{1em}
    \centering
    \scriptsize
    \setlength{\tabcolsep}{11pt}
    
    \begin{tabular}{l|cccccc}
    \toprule
    & \multicolumn{6}{c}{\textbf{Bias Type}}\\
        Model & Avg & Selection & Attrition & Performance & Detection & Reporting \\ 
    & & n = 157 & n = 46 & n= 54 & n = 61 & n = 14\\
    \midrule
OpenAI-v3 & 22.72 & 40.13 $\pm$ 3.56 & 6.3 $\pm$ 2.72 & 26.67 $\pm$ 5.29 & 27.38 $\pm$ 4.81 & 13.1 $\pm$ 7.56 \\
GritLM-7B & 18.87 & 35.84 $\pm$ 3.59 & 6.99 $\pm$ 2.67 & 15.19 $\pm$ 3.76 & 26.83 $\pm$ 5.08 & 9.52 $\pm$ 7.09 \\
\midrule
GPT-4o & \textbf{47.48}  & 60.5 $\pm$ 3.54 & \textbf{42.36 $\pm$ 6.47} & 41.48 $\pm$ 6.06 & 50.05 $\pm$ 5.7 & \textbf{43.03 $\pm$ 11.55} \\
Sonnet-3.5 & 39.17 & 55.42 $\pm$ 3.62 & 30.65 $\pm$ 5.71 & 37.01 $\pm$ 5.69 & 37.21 $\pm$ 5.5 & 35.54 $\pm$ 11.55 \\
Llama-3.1-70B & 45.62 & 61.17 $\pm$ 3.48 & 34.21 $\pm$ 5.95 & 45.19 $\pm$ 5.96 & \textbf{50.11 $\pm$ 5.68} & 37.41 $\pm$ 11.49 \\
Llama-3-8B & 22.66  & 49.44 $\pm$ 3.68 & 14.49 $\pm$ 4.86 & 22.13 $\pm$ 4.95 & 20.77 $\pm$ 4.7 & 6.46 $\pm$ 4.4 \\
\midrule
Llama-3-8B Fine-Tuned & 40.8  & \textbf{69.01 $\pm$ 3.41} & 24.78 $\pm$ 5.85 & \textbf{47.28 $\pm$ 6.26} & 48.63 $\pm$ 5.92 & 14.29 $\pm$ 9.35 \\

\bottomrule
\end{tabular}
\caption{{\small Results for Sentence Retrieval Standard Error Included}.}
\label{tab:task_2_sentence_retrieval_results_ste}
\end{table*}
\begin{table*}[htbp]
\vspace{1em}
    \centering
    \scriptsize
    \setlength{\tabcolsep}{11pt}
    
    \begin{tabular}{l|cccccc}
    \toprule
    & \multicolumn{6}{c}{\textbf{Bias Type}}\\
        Model & Full & Selection & Attrition & Performance & Detection & Reporting \\ 
    & n = 310 & n = 86 & n = 60 & n = 52 & n = 56& n = 50 \\
    \midrule
GPT-4o & 47.17  & 58.46 $\pm$ 4.37 & 60.2 $\pm$ 5.04 & 48.28 $\pm$ 5.24 & 42.68 $\pm$ 5.29 & 26.25 $\pm$ 4.7 \\
Sonnet-3.5 & \textbf{59.92} & \textbf{73.08 $\pm$ 3.89} & \textbf{73.47 $\pm$ 4.28} & \textbf{50.57 $\pm$ 5.42} & \textbf{51.22 $\pm$ 5.44} & \textbf{51.25 $\pm$ 5.26} \\
Llama-3.1-70B & 53.16  & 66.15 $\pm$ 3.89 & 62.24 $\pm$ 4.98 & 44.83 $\pm$ 5.19 & 46.34 $\pm$ 5.64 & 46.25 $\pm$ 5.7 \\
Llama-3-8B & 26.54  & 26.92 $\pm$ 3.87 & 34.69 $\pm$ 4.77 & 24.14 $\pm$ 4.64 & 21.95 $\pm$ 4.69 & 25.0 $\pm$ 4.91 \\
\midrule
Llama-3-8B Fine-Tuned & 29.6  & 40.77 $\pm$ 4.36 & 28.57 $\pm$ 4.37 & 24.14 $\pm$ 4.42 & 19.51 $\pm$ 4.52 & 35.0 $\pm$ 5.38 \\
\bottomrule
\end{tabular}%
\caption{{\small Results for Support Judgment Selection Standard Error Included}.}
\label{tab:task_3_support_judgment_selection_results_ste}
\end{table*}
\begin{table*}[htbp]
\vspace{1em}
    \centering
    \scriptsize
    \setlength{\tabcolsep}{7pt}
    
    \begin{tabular}{l|ccccccc}
    \toprule
    & \multicolumn{7}{c}{\textbf{Bias Type}}\\
    
    Model & Avg & Selection & Attrition & Performance & Detection & Reporting & Deviation\\ 
    &  &  n = 933 & n = 629 & n = 309 & n = 645 & n = 594 & n= 331\\
    \midrule
GPT-4o & \textbf{42.07}  & 50.52 $\pm$ 1.87 & 36.48 $\pm$ 2.14 & \textbf{54.67 $\pm$ 2.8} & 43.66 $\pm$ 2.02 & \textbf{34.43 $\pm$ 1.87} & 32.65 $\pm$ 2.77 \\
Sonnet-3.5 & 41.93  & \textbf{52.83 $\pm$ 1.86} & \textbf{37.07 $\pm$ 2.04} & 50.56 $\pm$ 2.75 & 43.29 $\pm$ 2.09 & 33.58 $\pm$ 1.41 & \textbf{34.23 $\pm$ 1.96} \\
Llama-3.1-70B & 38.81 & 48.1 $\pm$ 1.85 & 31.7 $\pm$ 1.94 & 48.04 $\pm$ 2.67 & \textbf{44.36 $\pm$ 2.0} & 31.58 $\pm$ 1.33 & 29.04 $\pm$ 2.37 \\
Llama-3-8B & 30.05 & 36.37 $\pm$ 1.28 & 32.44 $\pm$ 1.62 & 39.13 $\pm$ 2.59 & 37.21 $\pm$ 1.92 & 19.76 $\pm$ 1.96 & 15.38 $\pm$ 1.64 \\
\midrule
Llama-3-8B Fine-Tuned & 36.33  & 49.51 $\pm$ 1.91 & 33.34 $\pm$ 1.89 & 42.44 $\pm$ 2.69 & 40.1 $\pm$ 2.01 & 26.95 $\pm$ 0.44 & 25.67 $\pm$ 1.06 \\
\bottomrule
\end{tabular}
\caption{{\small Results for Risk Level Determination Standard Error Included}.}
\label{tab:task_4_risk_level_determination_results_ste}
\end{table*}

\twocolumn

\end{document}